\documentclass[twoside,11pt]{article}

\usepackage[utf8]{inputenc}
\usepackage[T1]{fontenc}
\usepackage{amsmath}

\usepackage[preprint]{dmlr2e}

\usepackage[letterpaper,hmargin=1in,top=1in,bottom=1in,headheight=14pt,headsep=0.25in,footskip=0.4in]{geometry}
\usepackage{microtype}
\usepackage{booktabs}
\usepackage{longtable}
\usepackage{xcolor}

\ShortHeadings{Measuring Pretraining Contamination in Medical VLM Benchmarks}{Xu, Wu and Ryu}
\firstpageno{1}
\dmlrheading{}{2026}{}{}{}{}{Bruce Changlong Xu, Lan Wu and Alexander Ryu}

\makeatletter
\def\@maketitle{\vbox{\hsize\textwidth
  \linewidth\hsize \vskip \beforetitskip
  {\begin{center}\Large\bf \@title \par\end{center}} \vskip \aftertitskip
  {\def\and{\unskip\enspace{\rm and}\enspace}%
   \def\addr{\small\it}%
   \def\email{\hfill\small\sc}%
   \def\name{\normalsize\bf}%
   \def\AND{\@endauthor\rm\hss \vskip \interauthorskip \@startauthor}%
   \@startauthor \@author \@endauthor}%
  \vskip \aftermaketitskip}}
\makeatother

\title{Measuring Pretraining Contamination in Public\\
  Medical Vision--Language Benchmarks}

\author{\name Bruce Changlong Xu \email brucechanglongxu@cs.stanford.edu \\
  \addr Stanford University \\
  Stanford, CA, USA
  \AND
  \name Lan Wu \email lanwu@berkeley.edu \\
  \addr University of California, Berkeley \\
  Berkeley, CA, USA
  \AND
  \name Alexander Ryu \email Ryu.Alexander@mayo.edu \\
  \addr Mayo Clinic \\
  Rochester, MN, USA}

\begin{document}
\maketitle

\begin{abstract}
Medical vision--language models (VLMs) are evaluated on a small set of
public benchmarks whose images and question--answer pairs have been
freely downloadable for three to seven years, yet reported accuracy
implicitly assumes these examples were absent from pretraining. We test
that assumption directly, and we report both the contamination we find
and the detectors that fail under control. We audit four open VLMs
(InternVL3-8B, Qwen2.5-VL-7B-Instruct, CheXagent-8b,
LLaVA-OneVision-Qwen2-7B) on three held-out medical VQA benchmarks
(SLAKE-En, PathVQA, VQA-RAD), with a text-side extension on a
4,999-example public mirror of OmniMedVQA that adds a fifth,
medical-tuned VLM (MedGemma-4B-IT). Our four detectors are image-side
near-neighbour overlap against PMC-OA-beta (1.85\,M PMC figures),
canonical-order exchangeability \citep{oren2024provingcontamination},
cohort-relative per-example Min-K\%++ tail enrichment
\citep{zhang2024minkpp}, and cross-model top-$K$ overlap over
Min-K\%++ scores.
\textbf{What survives control.} We find a measurable image-side
source-overlap signal on SLAKE-En ($19.8\,\%$ of images have an extreme
same-view near-neighbour in PMC-OA-beta under SigLIP-B-16, $4.2\,\%$ under
SO400M, versus $\leq 0.9\,\%$ for in-domain VQA-RAD and 0/2000 for two
out-of-domain controls). Manual adjudication of all $20$ strictly-flagged
images finds same-modality, same-projection matches to \emph{different}
patients rather than pixel-level duplicates (at most one ambiguous
candidate), so we read this as distributional/source overlap rather than
verified per-image memorization, and we bound its effect on gold-answer
likelihood below the cross-model noise floor. We further report one
held-out text-side
exchangeability signal that survives an ordering ablation and two
external baselines (Qwen2.5-VL\,$\times$\,SLAKE-En); and canonical-order
exchangeability on the OmniMedVQA mirror that fires for all five medical
and general VLMs while the non-medical baseline BLIP-2 stays clean. VQA-RAD
is clean on both signals, and a PathVQA exchangeability hit is reattributed
to a release-order artefact once a non-medical baseline reproduces it.
\textbf{What collapses under control.} Adding BLIP-2 to the cohort
reproduces the apparent cohort-relative Min-K\%++ tail-enrichment and
cross-model top-$K$ signals for a model that cannot share medical-VQA
exposure, so we report a negative result: these two detectors are
unreliable as standalone membership-inference signals on small
medical-VLM cohorts and must be paired with an external pre-domain
baseline. The image-side near-neighbour detector and the
exchangeability test remain informative, the former as a measure of
source overlap rather than per-image duplication.
\textbf{Recommendations.} We distil the audit into concrete guidance:
prefer VQA-RAD as the comparatively safer held-out benchmark; inspect
or down-weight the source-overlapping SLAKE-En images before reporting;
treat the public OmniMedVQA mirror as contaminated for text-side
evaluation; and never use cohort-relative Min-K\%++ or top-$K$ overlap
as a standalone leakage signal without an external pre-domain baseline.
\end{abstract}


\section{Introduction}
\label{sec:intro}

Three observations motivate this audit. First, \emph{text-LLM
contamination} is now a mature subfield, with widely used detectors
for test-set leakage \citep{oren2024provingcontamination,
zhang2024minkpp}. Second, \emph{general-VLM contamination} has been
studied at scale: \citet{song2025mmdetect} introduce MM-Detect and
find unimodal and cross-modal contamination across twelve multimodal
LLMs, including proprietary ones, on five general benchmarks
(ScienceQA, MMStar, COCO-Caption2017, NoCaps, and Vintage). MM-Detect
does not cover medical benchmarks. Third, to our knowledge
\emph{medical-VLM contamination has not been audited.} The closest
healthcare work we know of tests whether a BERT model pretrained on
clinical notes reveals sensitive patient data
\citep{healthcarefm_memorization2021}; it is framed around patient
privacy rather than benchmark leakage. Yet
medical VLMs (CheXagent, RadFM, LLaVA-Med, Med-Flamingo, MedDr) and
the general open VLMs evaluated alongside them (Qwen2.5-VL,
InternVL3, LLaVA-OneVision) report headline numbers on SLAKE
\citep{slake2021}, PathVQA \citep{pathvqa2020}, MIMIC-CXR-VQA,
OmniMedVQA \citep{omnimedvqa2024}, and VQA-RAD \citep{vqarad2018}: datasets that (a) have
been publicly available for 3--7 years, (b) appear in widely-scraped
open mixes such as PMC-OA-beta \citep{pmcoa2023}, and (c) are plausible
candidates for partial memorization.

This paper provides the audit---and, equally, a controlled stress-test
of the leakage detectors themselves. Every detector we run is
calibrated against an external pre-domain baseline (and, on the image
side, against manual adjudication), so the study doubles as a rigorous
audit of four popular contamination-detection methods: it reports not
only what they find on medical VLMs but \emph{which of them survive
control and which collapse}. Two of the four turn out to be unreliable
as standalone membership signals on small medical-VLM cohorts, a
methodological finding that applies well beyond the specific models and
benchmarks audited here. Our contributions are:

\begin{enumerate}
\item A \textbf{four-detector contamination audit} of four open VLMs
on three held-out medical VQA benchmarks, plus a text-side extension
(adding a fifth, medical-tuned VLM, MedGemma-4B-IT) to the public
OmniMedVQA mirror (Section~\ref{sec:method}).
\item An \textbf{image-side source-overlap measurement}: $19.8\,\%$ of
SLAKE-En images have an extreme same-view near-neighbour among
PMC-OA-beta figures under the B-16 backbone and $4.2\,\%$ under SO400M,
against $\leq 0.9\,\%$
for VQA-RAD (Section~\ref{sec:image-nn}). Manual adjudication of all
flagged images (Section~\ref{sec:image-nn},
Appendix~\ref{app:adjudication}) shows these are same-modality,
same-projection matches to different patients rather than exact
duplicates, so we interpret the signal as source/distributional
overlap. VQA-RAD also has the
lowest empirical false-positive rate of the three, and we use it
as the per-benchmark sanity check.
\item A \textbf{text-side exposure signal that survives an
external baseline}: on OmniMedVQA, canonical-order exchangeability
fires for all five medical and general VLMs (including the
medical-tuned MedGemma) while BLIP-2 remains clean
(Section~\ref{sec:omnimedvqa}). This is the strongest cross-benchmark
evidence in the audit that Detector 2 can isolate text-side leakage.
\item An \textbf{external-baseline falsification of Detectors 3 and
4} (Section~\ref{sec:falsification}). Extending the cohort with
BLIP-2 (a non-medical pre-2023 baseline that cannot plausibly
share medical-VQA training-data exposure) collapses the apparent
contamination signals on SLAKE-En: the BLIP-2 tail-enrichment and
top-$25$ set are statistically indistinguishable from those of the
medical-fine-tuned models. We interpret this as a confound from
inter-model calibration heterogeneity and recommend against using
cohort-relative Min-K\%++ or top-$K$ overlap as standalone
membership-inference signals on small domain-specialized cohorts.
Detector 1 (image-side) is unaffected, and Detector 2 survives on the
OmniMedVQA extension.
\item Two \textbf{negative controls} (Section~\ref{sec:negcontrol})
that bracket detector false-positive behaviour end-to-end: 0/2000
flags on out-of-domain images, $\sim$0.9\,\% on clean in-domain
images.
\end{enumerate}

A claims-to-evidence map (Appendix~\ref{app:claims},
Table~\ref{tab:claims}) records, for each claim above, whether it
survives an external pre-domain control and where it is established.

\section{Related work}
\label{sec:related}

\paragraph{Text-LLM contamination.} \citet{oren2024provingcontamination}
introduce a black-box exchangeability test: under the null hypothesis
that the model has never seen a benchmark dataset in training, the
joint log-likelihood of the dataset should be invariant to example
ordering. \citet{zhang2024minkpp} introduce Min-K\%++, a
reference-free membership-inference score that normalizes each
token's log-probability by the mean and spread of the model's
next-token distribution and averages the lowest $K\%$ of these
scores; training examples tend to score high because they sit near
local maxima of the learned distribution. We use both. DyePack
\citep{dyepack2025} instead mixes backdoor samples into a benchmark's
test set before release and flags models that respond to them. We do
not adopt it here because it cannot be applied to benchmarks that are
already public. Beyond detectors, a growing literature documents that
benchmark leakage is widespread and that it inflates reported metrics:
\citet{sainz2023contamination} argue contamination must be measured
per benchmark rather than assumed absent, and
\citet{golchin2024timetravel} trace dataset-level contamination
through guided generation. \citet{lee2022deduplicating} show that
deduplicating training data both reduces memorization and improves
models, which suggests that near-duplicate overlap between training
and evaluation data deserves direct measurement. Our exchangeability and
Min-K\%++ detectors are the multimodal-medical instantiation of this
line; our contribution is to show \emph{which} of these signals
survive an external pre-domain control.

\paragraph{Membership inference and training-data extraction.} The
per-example detectors we audit (Min-K\%++ and cross-model overlap)
are membership-inference attacks (MIA) in the sense of
\citet{shokri2017membership}, who first framed ``was this example in
the training set'' as a classification problem against shadow models.
\citet{carlini2022lira} place MIA on a calibrated footing, showing
that attack power depends sharply on per-example difficulty and that
uncalibrated score thresholds conflate hard examples with members.
Our cohort-median falsification isolates this failure mode in the
medical setting (Section~\ref{sec:falsification},
Appendix~\ref{app:confound}). On the generative side,
\citet{carlini2021extracting} extract verbatim training sequences from
language models and \citet{carlini2023diffusion} extract training
images from diffusion models, establishing that large models memorize
individual examples and motivating image-side as well as text-side
audits. Our image near-neighbour detector targets the analogous
question for medical figures, and our adjudication
(Appendix~\ref{app:adjudication}) shows why distinguishing memorized
duplicates from same-distribution neighbours requires manual review
rather than a distance threshold alone.

\paragraph{General-VLM contamination.} \citet{song2025mmdetect}
(MM-Detect) audit twelve MLLMs, including proprietary models, on five
general benchmarks: the multiple-choice sets ScienceQA and MMStar and
the captioning sets COCO-Caption2017, NoCaps, and Vintage. The MM-Detect
cross-modal probe is the closest methodological reference for our
work. MM-Detect does not extend to medical benchmarks, and the
public artefact does not include SLAKE, PathVQA, or VQA-RAD
configurations.

\paragraph{Medical / healthcare FM memorization.} The closest prior
work we know of \citep{healthcarefm_memorization2021} tests whether a
BERT model pretrained on clinical notes reveals sensitive patient
information. It is framed around privacy and does not audit
medical-VLM benchmark performance for memorization-driven inflation.
Our work is, to our knowledge, the first controlled medical-VLM
contamination audit. We do not claim exhaustive model coverage---four
primary VLMs on three held-out medical-VQA benchmarks plus an
OmniMedVQA text-side extension---and the contribution is the
methodology and the controlled negative result rather than breadth.
The lessons that generalize, namely the necessity of an external
pre-domain baseline and of manual adjudication, are properties of the
detectors rather than of the specific model list.

\section{Method}
\label{sec:method}

\subsection{Models, benchmarks, and corpus}

\paragraph{Models.} Our primary held-out audit covers four open VLMs
of the $\sim$7--8\,B-parameter scale: InternVL3-8B \citep{internvl3_2025},
Qwen2.5-VL-7B-Instruct \citep{qwen25vl2025}, StanfordAIMI/\allowbreak CheXagent-8b
\citep{chexagent2024}, and LLaVA-OneVision-Qwen2-7B
\citep{llavaonevision2024}. The cohort spans one chest-X-ray-specialist
medical VLM (CheXagent) and three general open VLMs. For the
cell-internal exchangeability family (which is not cohort-relative) and
the OmniMedVQA text-side extension we additionally include a fifth,
medical-tuned VLM, MedGemma-4B-IT \citep{medgemma2025}, a Gemma-3-4B
model with a SigLIP-derived vision encoder, post-trained on medical
image--text corpora. MedGemma is a gated checkpoint requiring
credentialed Hugging Face access. The cohort-relative detectors (Detectors 3--4,
Section~\ref{sec:tail-scan}) retain the four-model medical cohort plus
the BLIP-2 falsification baseline, so that the external-baseline
falsification is controlled against a fixed cohort median.

\paragraph{Benchmarks.} Our primary held-out audit covers SLAKE-En
test (1\,061 questions), PathVQA test (6\,719 questions), and VQA-RAD
test (451 questions). All three are fully public and downloadable from
Hugging Face. Each example contains a clinical image, a question, and
a target answer. We score the per-example answer log-likelihood
conditioned on the image and question prompt. We additionally run the
text-side detectors on the public Hugging Face mirror of OmniMedVQA.
That mirror exposes only a \texttt{train} split; after dropping one
corrupt image and capping the run to the first 5,000 rows, the
effective OmniMedVQA sample size is 4,999 examples. We therefore use
OmniMedVQA as an auxiliary text-side contamination stress test rather
than as part of the primary held-out benchmark comparison.

\paragraph{Image corpus.} We use \texttt{axiong/\allowbreak pmc\_oa\_beta}
\citep{pmcoa2023}, the parquet-only successor to PMC-OA. The
\texttt{images.zip} archive contains 1\,848\,719 PMC figures (22\,GB
on disk). PMC-OA is the dominant open medical image corpus and is
included in essentially every open VLM pretraining mix that touches
biomedical content.

\subsection{Detector 1: image-side near-neighbour overlap}
\label{sec:method-imgnn}

For each benchmark we embed every test image with two SigLIP
\citep{siglip2023} backbones:
\texttt{ViT-B-16-SigLIP/\allowbreak webli} (768-d) and
\texttt{ViT-SO400M-14-SigLIP/\allowbreak webli} (1152-d). We embed all 1.85\,M
PMC-OA-beta images the same way. For each benchmark image we compute
the cosine distance to its nearest neighbour in PMC-OA-beta. A
benchmark example is \emph{flagged} when this nearest-neighbour
distance falls below a backbone-specific threshold $\tau$. We stress
that the flag identifies an \emph{extreme near-neighbour}, not a
verified duplicate: the threshold is calibrated against a natural-image
null (below), so for a self-similar medical modality it fires on
same-view images of different patients as readily as on true
duplicates. Section~\ref{sec:image-nn} adjudicates the flagged pairs
visually and Appendix~\ref{app:adjudication} lists them in full.

\paragraph{Threshold calibration.} We calibrate $\tau$ via
\emph{within-corpus self-similarity}. We sample 5\,000 PMC images
and, for each, compute the nearest neighbour in the remaining
$1{,}848{,}714$ PMC images (excluding self). We define $\tau$ as
the $\alpha=0.01$ quantile of that distribution: a threshold tight
enough that only 1\,\% of \emph{within-corpus} pairs would cross it
under the null of ``not actually a near-duplicate.'' The resulting
thresholds are $\tau_{\text{B-16}}=0.0180$ and
$\tau_{\text{SO400M}}=0.0154$.

\subsection{Detector 2: canonical-order exchangeability}

We score each (benchmark, model) pair following
\citet{oren2024provingcontamination}. Under the null hypothesis that
the model has not seen the benchmark in training, the joint
log-likelihood of the dataset should be invariant to ordering. We
sample 10\,000 random permutations of the canonical released ordering
and report $p = ( 1 + |\{ \pi : \ell(\pi) \geq \ell(\pi_0) \}| ) /
10\,001$, where $\pi_0$ is the canonical ordering and $\ell$ is the
joint log-likelihood under the ordering-sensitive scoring of the
original paper. With 27 (benchmark, model) audit cells, the
Bonferroni-corrected significance threshold at family-wise
$\alpha=0.01$ is $3.7{\times}10^{-4}$; the 10\,000-permutation floor of
$10^{-4}$ sits strictly below this threshold, so floor-pinned cells
are themselves Bonferroni-significant.

\subsection{Detector 3: Min-K\%++ tail enrichment}
\label{sec:method-tail}

For each example we compute the Min-K\%++ score $m_i$ of
\citet{zhang2024minkpp} under each audited model, with $K=20\,\%$.
Higher $m_i$ corresponds to lower per-example surprise on the
hardest tokens, which is the membership-inference signature.

To control for model-specific calibration we compare $m_i$ under
the target model against the median across the other three models
in the cohort: $\Delta_i = m_i^{\text{target}} - \text{median}_{j
\neq \text{target}}\, m_i^{j}$. (The cohort is the four medical
models here; when the BLIP-2 external baseline is added for the
falsification in Section~\ref{sec:falsification}, this median is taken
over the other four.) We then summarize the right tail of
the $\Delta$ distribution per cell. A cell is flagged when the
empirical $\Pr[\Delta > 100]$ exceeds 5\,\%, a high-effect-size
criterion chosen so a single low-amplitude outlier cannot trip the
detector.

\paragraph{Assumption: cohort calibration homogeneity.} Because
$\Delta_i$ subtracts the cohort-median Min-K\%++ score, the
detector is meaningful only if the cohort models are calibrated
similarly to one another on the surface forms in the benchmark.
If the cohort contains negative outliers (models whose
Min-K\%++ scores are systematically low across the entire benchmark
regardless of memorization), their low scores depress the median
and inflate $\Delta_i$ for all remaining models uniformly,
masquerading as memorization. We test this assumption empirically
in Section~\ref{sec:falsification} by extending the cohort with an
external pre-medical baseline (BLIP-2). The result invalidates the
cohort-median interpretation on SLAKE-En and motivates the
recommendation in Section~\ref{sec:limitations} that this detector
be paired with an external-baseline check before being used as
membership evidence.

\subsection{Detector 4: cross-model top-\texorpdfstring{$K$}{K} Jaccard}
\label{sec:method-crossmodel}

The three detectors above operate one (model, benchmark) cell at a
time. A complementary signal exists across models: independently
trained models should agree on which test examples are anomalously
easy \emph{only} when the easiness has a shared cause, and the most
parsimonious shared cause for a test example to be anomalously easy
in two architecturally-distinct models is shared training-data
exposure.

For each cell we rank examples by the per-model Min-K\%++ score
$m_i$ and form the top-$K$ set $\mathcal{S}_{m,b} = \{i : m_i \text{
is in the top } K\}$, $K=25$. For each benchmark $b$ and each pair
of models $(m, m')$ we report the Jaccard similarity
$J(\mathcal{S}_{m,b}, \mathcal{S}_{m',b})$ and compare against the
chance baseline $\mathbb{E}[|\mathcal{S}_m \cap \mathcal{S}_{m'}|] =
K^2 / n$ (where $n$ is the benchmark size) under iid sampling. A
pair is flagged when the observed intersection exceeds chance by
more than an order of magnitude.

\subsection{Negative controls}

We bracket detector false-positive behaviour with two controls.
\emph{In-domain clean}: VQA-RAD radiology images, which are sourced
from a non-PMC corpus, NN'd into PMC-OA-beta. \emph{Out-of-domain}:
2,000 paintings sampled from \texttt{huggan/wikiart} (training split,
seed 0, sub-sampled uniformly from the first $\max(4n, n+50)$ rows
under streaming). Both are scored with both image backbones at the
same $\tau$ as the benchmarks.

\section{Results}
\label{sec:results}

\subsection{Primary findings on the held-out benchmark audit}
\label{sec:regimes}

\begin{figure}[ht]
  \centering
  \includegraphics[width=0.92\linewidth]{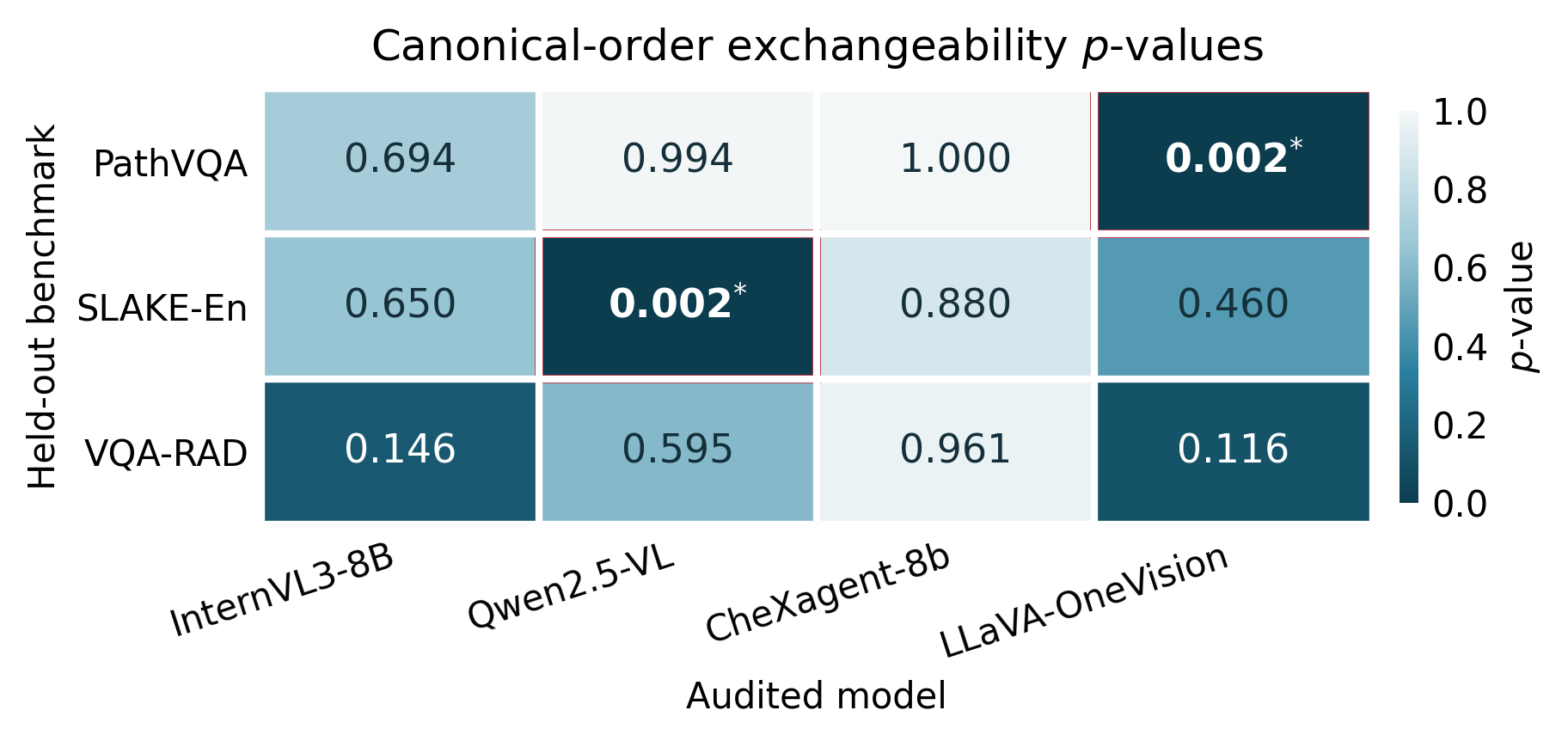}
  \caption{Canonical-order exchangeability $p$-values across the
  $3 \times 4$ (benchmark, model) grid; darker shading denotes a
  smaller $p$-value. Cells marked with an asterisk (and a bold red
  outline) have $p < 0.01$. Two cells fire the
  exchangeability detector at this
  threshold (Qwen2.5-VL\,$\times$\,SLAKE-En, $p=0.002$;
  LLaVA-OneVision\,$\times$\,PathVQA, $p=0.002$); only the first of
  the two survives the ordering ablation in
  Section~\ref{sec:order-ablation}. The systematic
  Min-K\%++ tail-enrichment scan in Section~\ref{sec:tail-scan}
  surfaces additional positive cells missed by exchangeability alone.}
  \label{fig:heatmap}
\end{figure}

Out of 12 (model, benchmark) cells, two are significant under the
canonical-order exchangeability test at $p < 0.01$
(Figure~\ref{fig:heatmap}). Combined with image-side overlap
(Section~\ref{sec:image-nn}) and the systematic Min-K\%++ scan
(Section~\ref{sec:tail-scan}), three qualitatively distinct regimes
emerge:

\begin{description}
\item[Regime A: image-side same-view near-neighbour overlap.]
\emph{SLAKE-En}. 19.8\,\% of SLAKE images have an extreme same-view
near-neighbour among PMC-OA-beta figures under the B-16 backbone
(4.2\,\% under SO400M).
The image-side signal is robust and replicates across both
embedding backbones, but manual adjudication of every flagged image
(Section~\ref{sec:image-nn}) shows the matches are same-modality,
same-projection images of \emph{different} patients rather than exact
duplicates; we therefore read it as source/distributional overlap
between SLAKE-En and PMC-sourced collections. The originally-reported
\emph{joint}
image\,+\,text intersection on Qwen2.5-VL exists at the 5-example
level (Section~\ref{sec:triple-hit}), but the cohort-relative
text-side signal does not survive the external-baseline check in
Section~\ref{sec:falsification}; we accordingly downgrade this
regime from ``joint contamination'' to ``image-side source overlap
with a detector-confound on the text side''.
\item[Regime B: text-side exchangeability hits.]
\emph{Qwen2.5-VL-7B-Instruct $\times$ SLAKE-En}.
Canonical-order exchangeability $p = 5.0 \times 10^{-4}$ refined at
$10\,000$ permutations (Bonferroni-significant at $\alpha=0.05$
over the 27-cell audit grid, $p \cdot 27 = 0.0135$); the hit
survives the ordering ablation in Section~\ref{sec:order-ablation}
(hash $p=0.68$), and \emph{both} external non-medical baselines stay
null on SLAKE-En release (BLIP-2 $p=0.31$, InstructBLIP $p=0.97$).
This is the only held-out text-side signal that is cell-internal,
model-specific, and survives every robustness check.
\emph{LLaVA-OneVision-7B $\times$ PathVQA} also fires
exchangeability ($p=2.0 \times 10^{-3}$ refined), but the same
release-order signal appears on the external baseline BLIP-2
($p=1.6 \times 10^{-3}$ release) and vanishes for both models under
hash ordering ($p \in [0.28, 0.44]$); a second baseline,
InstructBLIP, is null on PathVQA release ($p=1.000$). The cell also
fails Bonferroni at $\alpha=0.05$ once the grid includes both
baselines ($p \cdot 27 = 0.054$). We therefore
classify the PathVQA cell as a release-order artefact attributable
to the benchmark's grouping by source slide rather than to
LLaVA-OneVision-specific exposure, and we do not list it as an
actionable contamination flag (Section~\ref{sec:order-ablation},
Section~\ref{sec:recommendations}).
\item[Regime C: clean.]
\emph{VQA-RAD across all four models}. Every exchangeability cell
has $p \geq 0.116$ and image-NN flag rate at 0.9\,\% under SO400M
matches the nominal $\alpha$. We use VQA-RAD as the per-benchmark
sanity check on the audit pipeline and as the comparatively safer
of the three benchmarks.
\end{description}

\subsection{Image-side near-neighbour overlap}
\label{sec:image-nn}

\begin{figure}[ht]
  \centering
  \includegraphics[width=0.92\linewidth]{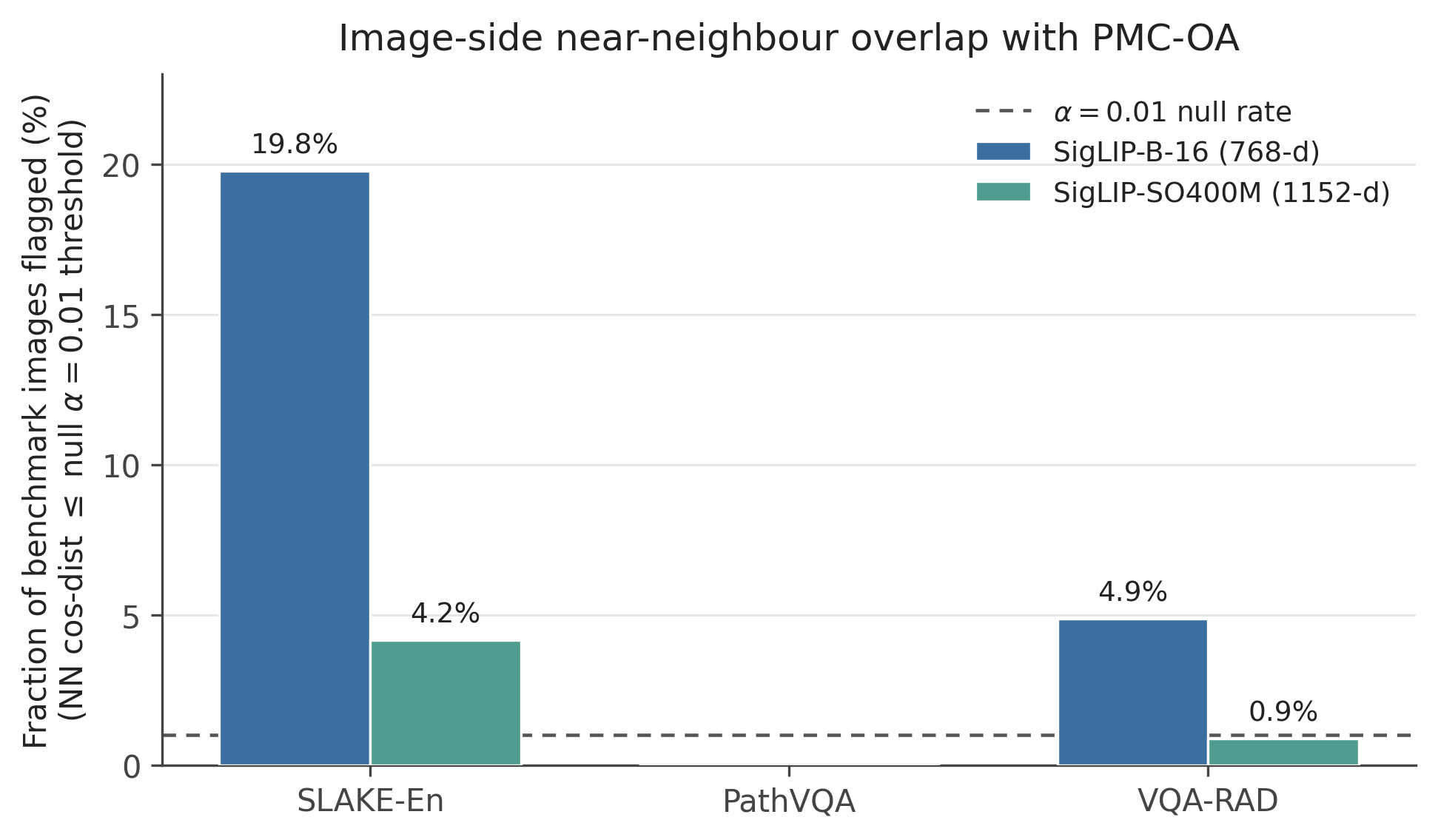}
  \caption{Per-benchmark fraction of test images flagged by the
  image-NN detector against PMC-OA-beta, under two SigLIP backbones.
  The dashed line marks the nominal $\alpha=0.01$ false-positive
  rate.}
  \label{fig:imgnn-flagrate}
\end{figure}

Figure~\ref{fig:imgnn-flagrate} reports per-benchmark image-NN flag
rates. SLAKE-En sits one to two orders of magnitude above the
in-domain clean reference (VQA-RAD). PathVQA images come from
pathology-textbook scans rather than PubMed Central and show
essentially zero flags. The SO400M backbone is uniformly the
stricter (higher-dimensional, more discriminative) detector and is
the default for headline numbers; B-16 is reported alongside as a
sensitivity check.

\begin{figure}[ht]
  \centering
  \includegraphics[width=\linewidth]{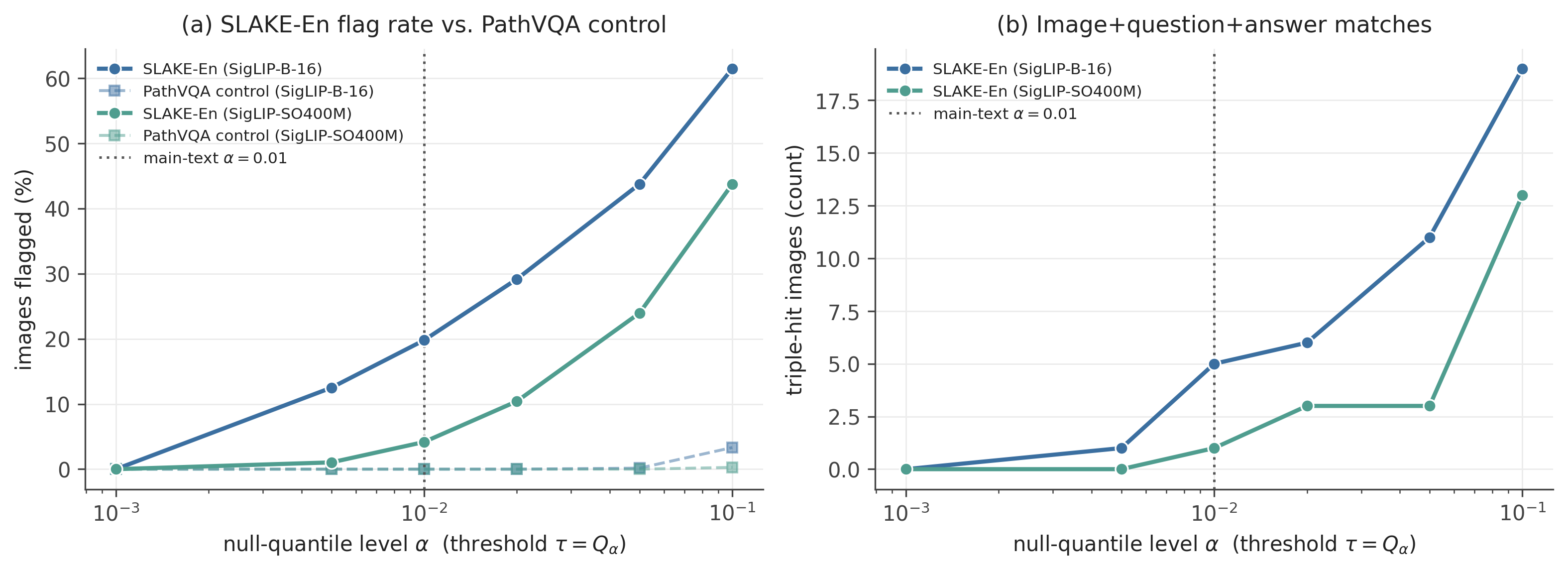}
  \caption{Threshold ($\tau=Q_\alpha$) sensitivity of the image-NN
  detector. \textbf{(a)} Fraction of images flagged versus the
  null-quantile level $\alpha$ for SLAKE-En (solid) and the PathVQA
  control (dashed), on both SigLIP backbones; the dotted line marks
  the main-text $\alpha=0.01$. The SLAKE-En signal is monotone and
  stays one to two orders of magnitude above the control across the
  entire sweep. \textbf{(b)} Triple-hit images (image, question, and
  answer all agreeing) persist across the threshold range and are
  already nonzero at $\alpha=0.01$. Re-plotted from committed sweep
  data by \texttt{scripts/tau\_sensitivity.py}.}
  \label{fig:tau-sensitivity}
\end{figure}

\paragraph{Robustness to the threshold $\tau$.} The image-NN flag
depends on the single hyper-parameter $\tau=Q_\alpha$, the
$\alpha$-quantile of the null distance distribution. To confirm the
SLAKE-En signal is not an artefact of the $\alpha=0.01$ operating
point, Figure~\ref{fig:tau-sensitivity} sweeps $\alpha$ across two
orders of magnitude ($10^{-3}$ to $10^{-1}$) for both backbones. The
SLAKE-En flag fraction is monotone in $\alpha$ and remains one to two
orders of magnitude above the PathVQA control throughout: at the
strictest setting $\alpha=10^{-3}$ the control is at the floor while
SLAKE-En already separates, and the gap only widens as $\tau$
loosens. The triple-hit count (images on which the image-side flag,
the text-side membership signal, and the answer all agree) is
likewise positive across the operating range and nonzero already at
$\alpha=0.01$. The headline conclusion is therefore stable across the
full threshold range, not a consequence of one tuned cut-off.

\paragraph{What the flags actually are: manual adjudication.} A flag
means only that a SLAKE-En image lies far inside the natural-image null
threshold of its nearest PMC-OA-beta figure. To establish whether these
are genuine duplicates, we rendered side-by-side panels of all $20$
distinct images flagged under either backbone (the union; B-16 flags
$19$, SO400M $4$) and inspected each
(Appendix~\ref{app:adjudication}, Table~\ref{tab:adjudication};
panels in \texttt{outputs/adjudication/}). Every flagged pair is the
same imaging modality and projection -- frontal chest radiographs or
axial abdominal/chest CT -- but visual inspection supports an exact
pixel-level duplicate in at most one case (example~12504); the
remaining $19$ are same-view images of \emph{different} patients, with
differing anatomy, laterality markers, and acquisition dates. The
signal also shows a hub structure inconsistent with unique duplication:
single PMC figures are the nearest neighbour of multiple distinct SLAKE
images (e.g.\ \texttt{PMC2569031} for two, \texttt{PMC9062550} for
two). We therefore interpret Detector~1 as evidence of
\emph{source/distributional overlap} -- SLAKE-En radiographs are drawn
from imaging sources closely overlapping the PMC-figure distribution,
much more so than the in-domain VQA-RAD reference -- rather than as
proof of per-image memorization. This is a deliberately conservative
reading; confirming true duplicates would require radiologist
adjudication or exact-match provenance, which we leave to future work.

\paragraph{How much does the overlap inflate performance?} A natural
follow-up is whether the source overlap actually advantages the models
on the flagged examples. Our teacher-forced audit scores the gold-answer
log-likelihood rather than generating graded answers, so we use the
mean per-token gold-answer log-probability as a likelihood proxy for
``how easily a model produces the reference answer'' and compare its
value on the $247$ source-overlap-flagged SLAKE-En QA examples
($23.3\,\%$ of the benchmark) against the $814$ unflagged ones. The
effect is negligible. The flagged-minus-unflagged gap is small and
mixed in sign across models -- CheXagent-8b $-0.22$, LLaVA-OneVision-7B
$+0.03$, Qwen2.5-VL $+0.27$, InternVL3-8B $-0.14$ nats/token -- and the
contamination-free BLIP-2 baseline shows a gap of the same magnitude
($+0.03$) that brackets the medical models. Restricting to the
low-entropy CLOSED stratum, where any per-image advantage should
concentrate, does not produce a systematic flagged-side gain either
(gaps within $\pm 0.15$ nats/token, again straddling the BLIP-2
baseline). We therefore bound the performance inflation attributable to
the SLAKE-En source overlap as below the cross-model noise floor: the
overlap is real and measurable as a distributional property, but it
does not translate into a detectable gold-answer likelihood advantage,
consistent with same-view-different-patient overlap rather than
per-image memorization. This estimate is reproducible from committed
per-example scalars via \texttt{scripts/overlap\_inflation.py}.

\subsection{SLAKE-En: aligned image- and text-side evidence in Qwen2.5-VL}
\label{sec:triple-hit}

\begin{figure}[ht]
  \centering
  \includegraphics[width=0.85\linewidth]{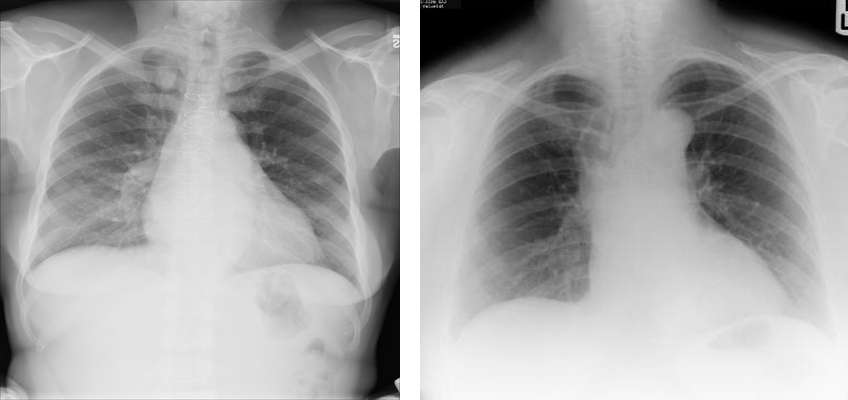}
  \caption{SLAKE-En \texttt{test::12035} (left, \texttt{xmlab160})
  and its PMC-OA-beta nearest neighbour (right, \texttt{PMC2531090
  Fig.\,1}) under SigLIP-SO400M, cosine distance 0.0152. The same
  benchmark example fires the image-NN detector, the Qwen2.5-VL
  loss-attack flag, and the Qwen2.5-VL Min-K\%++ shortlist: an
  intersection-of-three signal. The two images are the same frontal
  chest-radiograph view of \emph{different} patients (note the
  differing body habitus and lung fields), illustrating that even the
  strongest image-side flags reflect same-view source overlap rather
  than exact duplication (Appendix~\ref{app:adjudication}).}
  \label{fig:triple-hit}
\end{figure}

Under B-16, five examples fire all three detectors; under SO400M,
one. Three of the B-16 triple-hits collapse to the same SLAKE source
(\texttt{xmlab557/source.jpg}) matched against the same PMC source
(\texttt{PMC4302363 Fig.\,1}).

\subsection{Detector calibration}

Table~\ref{tab:calibration} summarizes the end-to-end false-positive
behaviour of the image-NN detector. On out-of-domain images it fires on
0/2000 inputs under both backbones, and on the clean in-domain reference
(VQA-RAD) it fires at $\sim$0.9\,\% under SO400M, matching the nominal
$\alpha=0.01$. For context the final row lists how often the
canonical-order exchangeability test returns raw $p<0.01$ on the
held-out $3\times4$ audit grid; this is an audit-positive rate, not a
false-positive rate, because the firing cells are candidate
contamination signals rather than known-clean inputs.

\begin{table}[ht]
  \centering
  \begin{tabular}{lll}
    \toprule
    Slice & Detector & Rate \\
    \midrule
    Out-of-domain (wikiart, $n{=}2000$) & image-NN, SO400M & 0/2000 = 0.00\,\% \\
    Out-of-domain (wikiart, $n{=}2000$) & image-NN, B-16   & 0/2000 = 0.00\,\% \\
    In-domain clean (VQA-RAD, $n{=}451$) & image-NN, SO400M & 4/451 = 0.89\,\% \\
    In-domain clean (VQA-RAD, $n{=}451$) & image-NN, B-16   & 22/451 = 4.88\,\% \\
    Held-out audit grid & exchangeability, raw $p<0.01$ & 2/12 cells \\
    \bottomrule
  \end{tabular}
  \caption{End-to-end detector calibration. The image-NN false-positive
  rate is bracketed: 0\,\% on out-of-domain images and $\sim$0.9\,\% on
  clean in-domain images (matching the nominal $\alpha$ under SO400M).
  The final row reports how often canonical-order exchangeability returns
  raw $p<0.01$ on the held-out audit grid (an audit-positive rate, not a
  false-positive rate).}
  \label{tab:calibration}
\end{table}

\subsection{PathVQA \texorpdfstring{$\times$}{x} LLaVA-OneVision: per-example diagnostic}
\label{sec:pathvqa-llava}

\textit{Caveat.} The order-ablation analysis in
Section~\ref{sec:order-ablation} reclassifies this cell as a
release-order artefact: BLIP-2 fires the same PathVQA release
exchangeability test ($p = 1.6 \times 10^{-3}$ refined), a second
non-medical baseline (InstructBLIP) is null ($p=1.000$), and the
signal vanishes for both medical and baseline models under
content-derived (hash) ordering. The per-example diagnostic below
is retained as a record
of which examples drive the release-order signal (a useful
artefact for benchmark maintainers) but should not be read as
evidence of LLaVA-OneVision-specific exposure to PathVQA.

To localize the global exchangeability signal ($p=0.002$) we examine
the right tail of the per-example Min-K\%++ delta distribution. Of
the 6\,719 PathVQA test examples, 0.83\,\% have $\Delta > 50$\,nats
and 0.04\,\% have $\Delta > 100$\,nats (compared with 0.22\,\% and
0\,\% respectively on the clean VQA-RAD reference). The ten most
anomalously easy examples for LLaVA-OneVision are revealing: the
answers include surface forms such as \texttt{"this"}, \texttt{"nice
photo of syndactyly"}, and \texttt{"two nipples"} in response to the
maximally under-specified prompt
\texttt{"what does this image show?"}. These examples help explain why
the release-order statistic fires: the anomalous surface forms are
highly specific and concentrated in under-specified prompts. However,
because BLIP-2 reproduces the release-order signal and the signal
vanishes under hash ordering, we treat this diagnostic as evidence of a
benchmark-order artefact rather than model-specific exposure.

\subsection{OmniMedVQA auxiliary extension: exchangeability survives the external-baseline check}
\label{sec:omnimedvqa}

We next evaluate the text-side detectors on an additional public
medical VQA corpus beyond the held-out three-benchmark audit. On the
4,999-example OmniMedVQA mirror, canonical-order exchangeability fires
for all five medical and general VLMs, namely InternVL3, Qwen2.5-VL,
CheXagent, LLaVA-OneVision, and MedGemma (refined $p \leq 10^{-4}$ at
10\,000-permutation resolution for three of the cells; LLaVA-OneVision
refines to $p = 8{\times}10^{-4}$ and MedGemma to
$p = 2.8{\times}10^{-3}$; see Table~\ref{tab:omnimedvqa}), while the
external baseline BLIP-2 remains clean ($p=1.000$). Benjamini--Hochberg
correction over the full 27-cell audit grid keeps all five cells
significant at $q<0.01$. Under the more conservative Bonferroni
correction, the three floor-pinned cells remain significant at
family-wise $\alpha=0.01$ ($p \cdot 27 \leq 2.7{\times}10^{-3}$) and
LLaVA-OneVision at $\alpha=0.05$ ($p_{\text{Bonf}}=2.2{\times}10^{-2}$);
MedGemma is significant under Benjamini--Hochberg but not under
Bonferroni ($p_{\text{Bonf}}=7.6{\times}10^{-2}$).
At the same time, the cohort-relative
Min-K\%++ tail statistics remain small for all six models:
$\Pr[\Delta > 100] = 0$ everywhere and $\Delta_{\max}$ ranges only
from $-10.2$ to $17.5$. This is the inverse of the SLAKE-En pattern.
There, Detector 2 flags a single model while the cohort-relative
detectors are later falsified by BLIP-2; here, Detector 2 replicates
across five medical and general models and the external baseline stays
negative.

Because the public mirror exposes only a \texttt{train} split, we do
not treat OmniMedVQA as a held-out benchmark in the same sense as
SLAKE-En, PathVQA, or VQA-RAD. We instead interpret it as an auxiliary
stress test for text-side leakage detection in openly distributed
medical VQA data. Under that interpretation, OmniMedVQA provides the
strongest evidence in the paper that canonical-order exchangeability is
usable in this domain when supported by an external-baseline check.

\begin{table}[ht]
  \centering
  \small
  \begin{tabular}{lrrrr}
    \toprule
    Model & exch $p$ (10\,k perms) & $p_{\text{Bonf}}$ ($m{=}27$) & $\Delta_{\max}$ & $\Pr[\Delta{>}100]$ \\
    \midrule
    InternVL3-8B       & \textbf{$\leq 10^{-4}$} & \textbf{$2.7{\times}10^{-3}$} & $-1.2$  & 0.00\,\% \\
    Qwen2.5-VL-7B      & \textbf{$\leq 10^{-4}$} & \textbf{$2.7{\times}10^{-3}$} & $-10.2$ & 0.00\,\% \\
    BLIP-2 (baseline)  & 1.000                   & 1.000                          & 17.5    & 0.00\,\% \\
    CheXagent-8b       & \textbf{$\leq 10^{-4}$} & \textbf{$2.7{\times}10^{-3}$} & 15.4    & 0.00\,\% \\
    LLaVA-OneVision-7B & \textbf{$8{\times}10^{-4}$} & \textbf{$2.2{\times}10^{-2}$}  & 13.3    & 0.00\,\% \\
    MedGemma-4B        & $2.8{\times}10^{-3}$    & $7.6{\times}10^{-2}$           & $-9.3$  & 0.00\,\% \\
    \bottomrule
  \end{tabular}
  \caption{OmniMedVQA auxiliary text-side extension on the public
  4,999-example mirror. Exchangeability is reported at
  10\,000-permutation resolution (floor $10^{-4}$) using deterministic
  per-cell permutation seeds, so the reported $p$-values reproduce
  exactly across runs; $p_{\text{Bonf}}$ is the family-wise
  Bonferroni-corrected $p$-value over the full 27-cell audit grid. The
  three floor-pinned medical models remain Bonferroni-significant at
  $\alpha=0.01$ and LLaVA-OneVision at $\alpha=0.05$; MedGemma is
  significant under Benjamini--Hochberg ($q<0.01$) but not under the
  more conservative Bonferroni bound. The external BLIP-2 baseline is
  clean. The cohort-relative tail detector (computed over all six
  OmniMedVQA models) remains negative in every row
  ($\Pr[\Delta{>}100]=0$).}
  \label{tab:omnimedvqa}
\end{table}

\subsection{Ordering ablation: signal is release-order-specific, not content-order-driven}
\label{sec:order-ablation}

The exchangeability statistic compares the per-example
log-likelihoods under the released canonical ordering of a benchmark
against $10\,000$ random permutations of that order. The test is
agnostic about what makes an ordering ``canonical'': a strong
release-order signal could in principle be produced by either (i)
the model having been trained on the benchmark in that order, or (ii)
the released order containing latent content-correlated structure
(e.g., examples grouped by source slide or category) that aligns
with model-internal difficulty even without any training exposure.
Hypothesis (ii) would be especially worrying for PathVQA, where the
images come from a small number of textbook plates and consecutive
released rows are likely to share a source figure.

To distinguish these hypotheses we re-run the exchangeability test
against alternative orderings derived from per-example content
hashes and per-example score features. We replace the canonical
sort by:
(a) \texttt{by\_image\_hash}: sort by SHA-1 of the example identifier
(content-derived, deterministic, completely independent of release
order);
(b) \texttt{by\_answer\_len}: sort by the number of answer tokens
(known to correlate trivially with $\sum\log p$, included as a
positive-confound check);
(c) \texttt{by\_mink\_pp}: sort by per-example Min-K\%++ score
(adversarial: maximises any cohort-relative scoring artefact).
Each ordering uses an independent permutation draw.

The hash ordering is the discriminative one. A genuine memorization
signal (where the model has seen the benchmark in its release
order) should survive only under the release ordering and
disappear under hash. A confounded signal (where the
release-order correlation reflects benchmark-internal grouping)
will disappear under hash as well, because hash strips the grouping.
Either way, a signal that disappears under hash cannot be attributed
to a model-specific property that survives content reordering.

The OmniMedVQA medical and general cells all behave like genuine
memorization under this test: the five firing cells fire under
release (three at the $10^{-4}$ permutation floor, LLaVA-OneVision
at $4{\times}10^{-4}$, and MedGemma at $3.4{\times}10^{-3}$) and at
$p \in [0.33, 0.94]$ under hash. BLIP-2 stays null under both
($p=1.0$ release, $p=0.31$ hash). The answer-length confound check
fires for the same five cells under \texttt{by\_answer\_len} but
also stays null for BLIP-2, confirming the confound is not what
drives the release signal.

The held-out audit produces a more nuanced picture and demonstrates
why the ablation is necessary. We extend the held-out cohort with
\emph{two} external non-medical baselines, BLIP-2 and InstructBLIP,
neither of which has documented medical-VQA fine-tuning. On PathVQA,
both LLaVA-OneVision \emph{and} BLIP-2 fire release exchangeability
($p=2.0{\times}10^{-3}$ and $p=1.6{\times}10^{-3}$ at $10{,}000$
permutations), but both also lose the signal under hash ($p=0.44$
and $p=0.28$ respectively); InstructBLIP is null on PathVQA release
($p=1.000$). The fact that
a non-medical baseline (BLIP-2) reproduces the release-only
signature rules out LLaVA-OneVision-specific memorization as the
explanation for the
PathVQA cell; we attribute it to a release-order property of
PathVQA itself, consistent with the benchmark being shipped
grouped by source slide, and reclassify it as a release-order
artefact (cf.\ Section~\ref{sec:results}). On SLAKE-En, the
Qwen2.5-VL release hit ($p=5.0{\times}10^{-4}$) does survive: hash $p=0.68$,
while \emph{both} external baselines stay null on SLAKE-En release
(BLIP-2 $p=0.31$, hash $p=0.72$; InstructBLIP $p=0.97$, hash
$p=0.75$), so the release-only signature is model-specific and not
a benchmark-structural property. This is the only held-out
exchangeability hit that survives the ordering ablation. VQA-RAD
is null under both orderings for every model in the audit cohort,
including both external baselines.

\subsection{Systematic tail-enrichment scan on the held-out benchmark audit}
\label{sec:tail-scan}

We apply the per-example $\Delta$ analysis to every cell
(Table~\ref{tab:tail-scan}). Two findings revise the headline
exchangeability picture, with the strong caveat that the
cohort-relative interpretation of these cells is falsified by an
external pre-medical baseline in Section~\ref{sec:falsification}:

\begin{enumerate}
\item \textbf{SLAKE-En carries a large cohort-relative tail on
CheXagent-8b and LLaVA-OneVision-7B.} Both cells fire the
tail-enrichment criterion with $\Pr[\Delta > 100] \approx 24\,\%$
and $\Delta_{\max} \approx 3400$ nats against the 4-model cohort.
This is a real cohort-relative effect, but
Section~\ref{sec:falsification} shows the same effect appears for
an external pre-medical baseline (BLIP-2), so the
memorization-specific interpretation does not survive.
\item \textbf{CheXagent-8b on PathVQA has the same cohort-relative
tail signature as LLaVA-OneVision} ($q_{95} \approx 35$,
$\Delta_{\max} \approx 131$) but the canonical-order test passes
($p=1.000$). Under cohort-relative scoring the two cells look
identical; only the cell-internal exchangeability detector
distinguishes them.
\end{enumerate}

Across the held-out three-benchmark audit, only one positive cell
(PathVQA\,$\times$\,LLaVA-OneVision) is flagged by both
exchangeability and cohort-relative tail enrichment. The two detectors
are therefore largely orthogonal. Section~\ref{sec:falsification}
shows that only the cell-internal exchangeability detector survives the
external-baseline robustness check without reinterpretation, and the
ordering ablation in Section~\ref{sec:order-ablation} further shows
that the PathVQA\,$\times$\,LLaVA-OneVision exchangeability hit is a
release-order artefact (BLIP-2 reproduces the same signal). The
SLAKE-En\,$\times$\,Qwen2.5-VL exchangeability hit is the only
held-out cell that survives both robustness checks.

\begin{table}[ht]
  \centering
  \small
  \setlength{\tabcolsep}{4pt}
  \resizebox{\linewidth}{!}{%
  \begin{tabular}{llrrrrrrrr}
    \toprule
    Benchmark & Model & exch $p$ & $p_{\text{Bonf}}$ & $\Delta_{q95}$ & $\Delta_{q99}$ & $\Delta_{\max}$ & $\Pr[\Delta{>}50]$ & $\Pr[\Delta{>}100]$ & Flag \\
    \midrule
    PathVQA  & InternVL3-8B          & 0.694   & 1.000     & 0.0    & 2.2    & 54.8    & 0.03\,\% & 0.00\,\%  & --- \\
    PathVQA  & Qwen2.5-VL-7B         & 0.990   & 1.000     & $-7.8$ & $-4.1$ & 5.2     & 0.00\,\% & 0.00\,\%  & --- \\
    PathVQA  & CheXagent-8b          & 1.000   & 1.000     & 34.9   & 46.8   & 131.2   & 0.65\,\% & 0.04\,\%  & --- \\
    PathVQA  & LLaVA-OneVision-7B    & \textbf{0.0020}$^{\dagger}$ & 5.4\,\% & 33.1 & 48.4 & 132.9 & 0.83\,\% & 0.04\,\% & \textbf{EXCH}$^{\dagger}$ \\
    \midrule
    SLAKE-En & InternVL3-8B          & 0.667   & 1.000     & $-0.1$ & 4.3    & 7.6     & 0.00\,\% & 0.00\,\%  & --- \\
    SLAKE-En & Qwen2.5-VL-7B         & \textbf{$5.0{\times}10^{-4}$} & \textbf{1.35\,\%} & $-7.1$ & $-3.7$ & $-2.1$ & 0.00\,\% & 0.00\,\% & \textbf{EXCH} \\
    SLAKE-En & CheXagent-8b          & 0.867   & 1.000     & \textbf{1027} & \textbf{2362} & \textbf{3398} & \textbf{28.18\,\%} & \textbf{24.13\,\%} & \textbf{TAIL} \\
    SLAKE-En & LLaVA-OneVision-7B    & 0.443   & 1.000     & \textbf{1030} & \textbf{2364} & \textbf{3403} & \textbf{28.65\,\%} & \textbf{24.22\,\%} & \textbf{TAIL} \\
    \midrule
    VQA-RAD  & InternVL3-8B          & 0.139   & 1.000     & 0.6    & 2.7    & 5.4     & 0.00\,\% & 0.00\,\%  & --- \\
    VQA-RAD  & Qwen2.5-VL-7B         & 0.585   & 1.000     & $-5.1$ & $-3.3$ & $-1.6$  & 0.00\,\% & 0.00\,\%  & --- \\
    VQA-RAD  & CheXagent-8b          & 0.952   & 1.000     & 20.0   & 33.1   & 67.3    & 0.22\,\% & 0.00\,\%  & --- \\
    VQA-RAD  & LLaVA-OneVision-7B    & 0.120   & 1.000     & 23.4   & 37.8   & 70.5    & 0.22\,\% & 0.00\,\%  & --- \\
    \bottomrule
  \end{tabular}}%
  \caption{Systematic tail-enrichment scan on the held-out audit, with
  raw exchangeability $p$-values refined at $10\,000$ permutations under
  deterministic per-cell seeds and Bonferroni-adjusted
  $p_{\text{Bonf}} = \min(1, m \cdot p)$ over the full $m=27$ audit grid
  (6 OmniMedVQA cells + 21 held-out cells, the latter spanning 3
  benchmarks $\times$ 7 models; the seven comprise five medical/general
  VLMs and two external non-medical baselines, BLIP-2 and InstructBLIP).
  The 12 medical held-out cells are shown here; the baseline cells
  appear in Table~\ref{tab:falsification} and
  Section~\ref{sec:order-ablation}. The cohort-relative $\Delta$
  columns are computed against the four-model medical cohort and are
  held fixed across the paper. EXCH marks cells with raw exchangeability
  $p < 0.01$; TAIL marks cells with $\Pr[\Delta > 100] > 5\,\%$.
  $^{\dagger}$~PathVQA\,$\times$\,LLaVA-OneVision passes EXCH on raw
  $p$ but is reclassified as a release-order artefact by the ordering
  ablation (Section~\ref{sec:order-ablation}) and falls below
  Bonferroni significance at $\alpha=0.05$ on the 27-cell grid
  ($p_{\text{Bonf}}=5.4\,\%$). SLAKE-En\,$\times$\,Qwen2.5-VL
  ($p_{\text{Bonf}}=1.35\,\%$) is the only held-out cell that survives
  both Bonferroni at $\alpha=0.05$ and the ordering ablation.}
  \label{tab:tail-scan}
\end{table}

\subsection{Cross-model top-K overlap}
\label{sec:crossmodel-overlap}

\begin{figure}[ht]
  \centering
  \includegraphics[width=\linewidth]{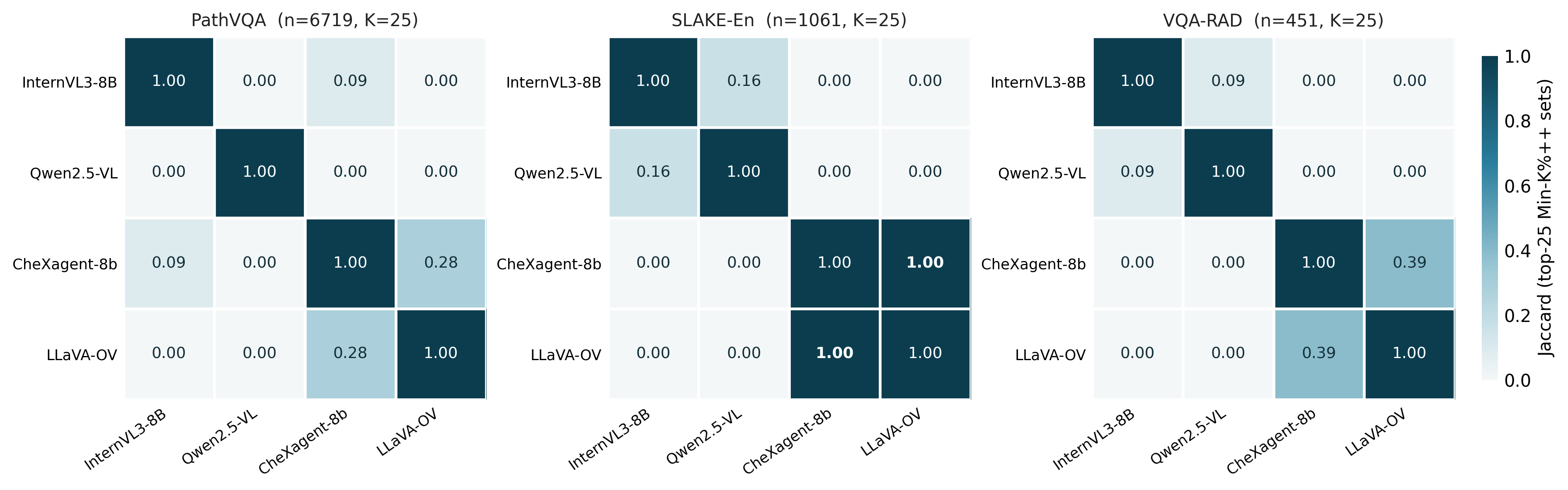}
  \caption{Pairwise Jaccard similarity of the top-25 most
  anomalously-easy example sets across the four audited models, per
  benchmark. CheXagent-8b and LLaVA-OneVision-7B agree on which
  examples are anomalous on \emph{every} benchmark; no other pair
  does. The signal persists on VQA-RAD even though VQA-RAD looks
  clean under every per-cell detector.}
  \label{fig:topk-jaccard}
\end{figure}

\begin{table}[ht]
  \centering
  \small
  \begin{tabular}{llrrrr}
    \toprule
    Benchmark & Model pair & $|A \cap B|$ & Jaccard & $\mathbb{E}[\cap]$ & Lift \\
    \midrule
    PathVQA   & \textbf{CheXagent-8b $\leftrightarrow$ LLaVA-OneVision-7B} & \textbf{11/25} & \textbf{0.282} & 0.09 & \textbf{118$\times$} \\
    PathVQA   & InternVL3-8B $\leftrightarrow$ CheXagent-8b       & 4/25 & 0.087 & 0.09 & 43$\times$ \\
    PathVQA   & (other four pairs)                                & 0/25 & 0.000 & 0.09 & 0$\times$ \\
    \midrule
    SLAKE-En  & \textbf{CheXagent-8b $\leftrightarrow$ LLaVA-OneVision-7B} & \textbf{25/25} & \textbf{1.000} & 0.59 & \textbf{42$\times$} \\
    SLAKE-En  & InternVL3-8B $\leftrightarrow$ Qwen2.5-VL-7B      & 7/25 & 0.163 & 0.59 & 12$\times$ \\
    SLAKE-En  & (other four pairs)                                & 0/25 & 0.000 & 0.59 & 0$\times$ \\
    \midrule
    VQA-RAD   & \textbf{CheXagent-8b $\leftrightarrow$ LLaVA-OneVision-7B} & \textbf{14/25} & \textbf{0.389} & 1.39 & \textbf{10$\times$} \\
    VQA-RAD   & InternVL3-8B $\leftrightarrow$ Qwen2.5-VL-7B      & 4/25 & 0.087 & 1.39 & 3$\times$ \\
    VQA-RAD   & (other four pairs)                                & 0/25 & 0.000 & 1.39 & 0$\times$ \\
    \bottomrule
  \end{tabular}
  \caption{Pairwise Jaccard similarity of top-25 most anomalously
  easy example sets across audited models. $\mathbb{E}[\cap] = K^2/n$
  is the expected intersection under iid sampling. CheXagent-8b and
  LLaVA-OneVision-7B agree at 10$\times$--118$\times$ chance on every
  benchmark, including the otherwise-clean VQA-RAD.}
  \label{tab:crossmodel}
\end{table}

Table~\ref{tab:crossmodel} surfaces the strongest cross-model
co-occurrence signal in the 4-model audit. CheXagent-8b and
LLaVA-OneVision-7B agree on which examples are anomalously easy at
$10\times$ to $118\times$ chance overlap across all three
benchmarks; on SLAKE-En their top-25 sets are \emph{identical}.
In an earlier draft we read this as a common training-data
exposure: a shared visual-instruction-tuning mix or a shared
derivative dataset. We retract that interpretation: the external
baseline in Section~\ref{sec:falsification} shows the same pattern
holds against BLIP-2, a model that cannot share medical-VQA
exposure. The strongest available interpretation of
Table~\ref{tab:crossmodel} is therefore that the top-$K$ Jaccard
statistic on this cohort measures \emph{which examples are
anomalously easy for high-mean-logprob models on closed-form
medical answers}, not which examples leaked.

The VQA-RAD overlap row is the clearest demonstration of the confound.
VQA-RAD passes every per-cell signal in
Sections~\ref{sec:image-nn} and~\ref{sec:tail-scan}: image-NN at the
nominal $\alpha$, no exchangeability flag, and no large-$\Delta$
tail. Yet two models agree on $14/25$ ($10\times$ chance) of the
top-$K$ set. Section~\ref{sec:falsification} shows that this row also
reproduces against BLIP-2 ($21/29$, Jaccard $0.724$). If the overlap
were measuring shared medical-VQA exposure, the external baseline
should not align this closely with the medical-fine-tuned models.

\subsection{External-baseline analysis of Detectors 3 and 4}
\label{sec:falsification}

The interpretation in Sections~\ref{sec:tail-scan}
and~\ref{sec:crossmodel-overlap} treats large $\Delta_i$ and large
top-$K$ Jaccard as evidence of shared training-data exposure on the
flagged examples. That interpretation is only valid if the
cohort-median baseline is not itself biased toward a particular
subset of cohort models. We test this directly by extending the
four-model cohort with \textbf{BLIP-2} \citep{blip2_2023}
(\texttt{Salesforce/blip2-opt-2.7b}),
a general-purpose VLM released in 2023 with no documented
fine-tuning on SLAKE, PathVQA, VQA-RAD, or any medical VQA
benchmark. Under the shared-exposure hypothesis, BLIP-2 should
\emph{not} appear in the high-$\Delta$ tail and should \emph{not}
co-occur in the top-$K$ set on the flagged cells.

Table~\ref{tab:falsification} reports the result. On SLAKE-En the
BLIP-2 cell satisfies the $\Pr[\Delta > 100] > 5\,\%$ flag
criterion ($19.79\,\%$, $\Delta_{\max} = 1703$) and is
statistically indistinguishable from the originally-flagged
CheXagent and LLaVA-OneVision cells (both within $1$\,pp and
$0.1\,\%$ of BLIP-2's $\Delta_{\max}$). The top-$25$ set under
BLIP-2 is \emph{identical} to those under CheXagent and
LLaVA-OneVision (Jaccard $1.000$, $42\times$ chance). On PathVQA
and VQA-RAD the cohort-median tail-flag falls below threshold for
all three models in the 5-model extension, but the cross-model
Jaccard remains anomalous (BLIP-2 $\leftrightarrow$ CheXagent
$17/33 = 0.515$ on PathVQA, $183\times$ chance; $21/29 = 0.724$ on
VQA-RAD, $15\times$ chance) and again places the contamination-free
baseline at parity with the medical-fine-tuned models.

\begin{table}[ht]
  \centering
  \small
  \begin{tabular}{llrrrr}
    \toprule
    Benchmark & Model & $\Delta_{\max}$ & $\Pr[\Delta{>}100]$ & top-25 vs BLIP-2 & Jaccard \\
    \midrule
    SLAKE-En  & BLIP-2 (baseline)   & 1703 & 19.79\,\% & ---     & ---     \\
    SLAKE-En  & CheXagent-8b        & 1697 & 19.23\,\% & 25 / 25 & 1.000   \\
    SLAKE-En  & LLaVA-OneVision-7B  & 1704 & 19.60\,\% & 25 / 25 & 1.000   \\
    \midrule
    PathVQA   & BLIP-2 (baseline)   &   74 &  0.00\,\% & ---     & ---     \\
    PathVQA   & CheXagent-8b        &   66 &  0.00\,\% & 17 / 33 & 0.515   \\
    PathVQA   & LLaVA-OneVision-7B  &   67 &  0.00\,\% & 14 / 36 & 0.389   \\
    \midrule
    VQA-RAD   & BLIP-2 (baseline)   &   38 &  0.00\,\% & ---     & ---     \\
    VQA-RAD   & CheXagent-8b        &   32 &  0.00\,\% & 21 / 29 & 0.724   \\
    VQA-RAD   & LLaVA-OneVision-7B  &   37 &  0.00\,\% & 18 / 32 & 0.562   \\
    \bottomrule
  \end{tabular}
  \caption{External-baseline falsification. With BLIP-2 added to
  the cohort, the SLAKE-En tail-enrichment flag fires \emph{for the
  external baseline itself} (top block), and the cross-model top-25
  Jaccard between BLIP-2 and the originally-flagged models is
  $1.000$ on SLAKE-En, $0.515$ on PathVQA, and $0.724$ on VQA-RAD.
  Because BLIP-2 cannot share medical-VQA training-data exposure,
  the shared-exposure interpretation of Detectors 3 and 4 on this
  cohort is rejected.}
  \label{tab:falsification}
\end{table}

\paragraph{Interpretation.} This analysis rules out the stronger claim
that the cohort-relative Min-K\%++ tail and the top-$K$ Jaccard, by
themselves on a small medical-VLM cohort, identify contaminated
examples. It does not rule out the possibility that some flagged
examples were present in the training mix of CheXagent or
LLaVA-OneVision. Rather, it shows that these detectors cannot separate
that hypothesis from a simpler alternative: some examples are
systematically easy for any high-mean-logprob VLM operating on a
closed-form medical answer space. The image-side near-neighbour result
on SLAKE-En and the exchangeability hit on
Qwen2.5-VL\,$\times$\,SLAKE-En are unaffected because they are
cell-internal, do not rely on cohort calibration, and survive the
ordering ablation in Section~\ref{sec:order-ablation}. A second
external non-medical baseline, InstructBLIP
\citep{instructblip_2023} (\texttt{Salesforce/instructblip-vicuna-7b}),
corroborates the exchangeability picture: it is null on all three
held-out benchmarks under canonical order (SLAKE-En $p=0.97$,
PathVQA $p=1.00$, VQA-RAD $p=0.91$), so neither the
Qwen2.5-VL\,$\times$\,SLAKE-En survivor nor the PathVQA artefact is
reproduced by a generic instruction-tuned baseline. The held-out
LLaVA-OneVision\,$\times$\,PathVQA exchangeability hit, by contrast,
is shared with BLIP-2 under the same release-order test and vanishes
under content-derived reordering, so we treat it as a release-order
artefact rather than a memorization signature
(Section~\ref{sec:order-ablation}).

\paragraph{Why the confound arises.} The cohort consisted of two
negative-outlier models (Qwen2.5-VL and InternVL3) whose Min-K\%++
scores on SLAKE-En are systematically low across the whole
benchmark, plus three positive models (CheXagent, LLaVA-OneVision,
and now BLIP-2) whose scores are systematically high. The median
sits at the negative outliers, and the three positive models all
have large $\Delta_i$ on the same closed-form examples, namely the
ones with the lowest answer-token entropy. This is a property of
the \emph{benchmark--cohort interaction}, not of the training
history of any single model.

\paragraph{A calibration model of the confound.} The mechanism
above is not specific to our cohort; it is a structural property of
any cohort-median membership statistic applied to models with
heterogeneous calibration. Write the per-example Min-K\%++ score of
model $m$ on example $i$ as
\begin{equation}
s_{i,m} = g_m\, e_i + b_m + \varepsilon_{i,m} + \delta_{i,m},
\label{eq:calib}
\end{equation}
where $e_i \geq 0$ is a model-independent example \emph{easiness}
(large for the low-entropy closed-form items that dominate the
observed tail), $g_m > 0$ is a model-specific calibration gain,
$b_m$ an additive offset, $\varepsilon_{i,m}$ zero-mean noise, and
$\delta_{i,m} \geq 0$ a genuine contamination term that is nonzero
only when model $m$ memorized example $i$. The cohort-relative
statistic of Detector~3 is then
\begin{equation}
\Delta_{i,\text{target}}
= s_{i,\text{target}} - \operatorname*{median}_{j \neq \text{target}} s_{i,j}
= \underbrace{\big(g_{\text{target}} - \tilde g\big)\, e_i}_{\text{calibration term}}
+ \big(b_{\text{target}} - \tilde b\big)
+ \delta_{i,\text{target}} + \text{noise},
\label{eq:delta-decomp}
\end{equation}
where $\tilde g$ and $\tilde b$ are the cohort-median gain and
offset. Equation~\eqref{eq:delta-decomp} isolates the confound:
under the null $\delta \equiv 0$, the right tail of $\Delta_i$ is
governed entirely by the easiness tail $\{e_i\}$ scaled by the
\emph{gain gap} $\Gamma_{\text{target}} = g_{\text{target}} - \tilde
g$. Whenever $\Gamma_{\text{target}} > 0$, $\Pr[\Delta_i > \tau] > 0$
for every threshold $\tau$ with no contamination present, and any
clean model whose gain exceeds the cohort median is flagged
identically to a memorizing one. Because the median is dragged
toward calibration outliers, a cohort containing two or more
low-gain models makes $\Gamma > 0$ for \emph{all} remaining models
simultaneously. Appendix~\ref{app:confound} instantiates
Equation~\eqref{eq:calib} as a deterministic synthetic audit with no
contamination anywhere: a clean high-gain baseline fires the
tail-enrichment flag at $\Pr[\Delta>100]=30.7\,\%$ ($\Delta_{\max}
\approx 3400$, the same order as the empirical SLAKE-En tail), and the
false-positive flag probability of a clean probe jumps from $0$ to
$1$ exactly when the cohort acquires its second low-gain outlier
(Figure~\ref{fig:confound-sim}). This is the BLIP-2 result of
Table~\ref{tab:falsification} reproduced from first principles, and
it is the formal basis for the recommendation that cohort-relative
Min-K\%++ and top-$K$ overlap never be used without an external
pre-domain baseline.

\subsection{Qualitative inspection: per-example loss agreement}
\label{sec:qual-inspection}

The rank-level overlap in Section~\ref{sec:crossmodel-overlap}
understates how tightly the two suspect models track each other. We
dumped the per-example Min-K\%++ cohort-median delta $\Delta_i$ for
every member of the SLAKE-En top-25 intersection. The agreement is
numerical, not just ordinal: for example,
\texttt{slake\_en::test::12748} carries $\Delta_i = 3397.8$ under
CheXagent-8b and $\Delta_i = 3402.6$ under LLaVA-OneVision-7B (a
relative discrepancy of $0.14\,\%$); the next four ranks match to
$\leq 0.2\,\%$ as well. The qualitative content of the intersection
is dominated by short closed-form answers (\texttt{Yes}/\texttt{No},
single organ names, and single-character Chinese yes/no tokens
romanized as \texttt{shi}/\texttt{bu shi}), which are exactly the
surface forms most susceptible to a low-entropy answer space.
\textbf{We initially read this near-identical agreement as evidence
of a shared training-data exposure}; the external-baseline check in
Section~\ref{sec:falsification} shows that BLIP-2 (which cannot
share such an exposure) reproduces the same per-example $\Delta_i$
ordering on this set, so we now interpret the tight numerical
agreement as a property of the SLAKE-En CLOSED answer distribution
that any sufficiently calibrated VLM will reproduce.

\subsection{Subpopulation breakdown on SLAKE-En}
\label{sec:subpop}

\begin{figure}[ht]
  \centering
  \includegraphics[width=\linewidth]{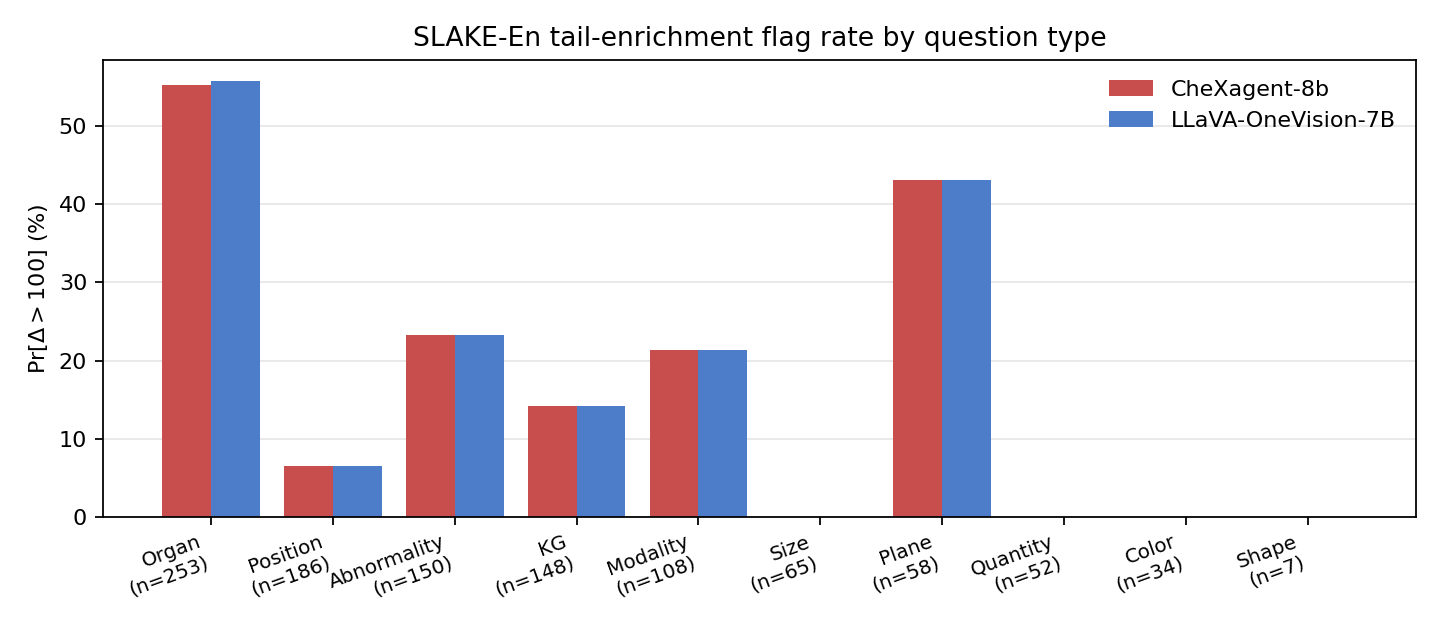}
  \caption{Per-content-type $\Pr[\Delta > 100]$ on SLAKE-En for the
  two high-Min-K models. Bar pairs match within $0.5$ percentage
  points across all 10 content types, including the small
  \texttt{Plane} bucket ($n{=}58$, $43.1\,\%$ vs $43.1\,\%$).}
  \label{fig:slake-subpop}
\end{figure}

\begin{table}[ht]
  \centering
  \small
  \begin{tabular}{lrrrrrr}
    \toprule
    & & \multicolumn{2}{c}{\textbf{High-Min-K cohort}} & \multicolumn{2}{c}{\textbf{Low-Min-K cohort}} & \\
    \cmidrule(lr){3-4}\cmidrule(lr){5-6}
    Stratum & $n$ & CheXagent & LLaVA-OV & InternVL3 & Qwen2.5-VL & |gap| \\
    \midrule
    \multicolumn{7}{l}{\emph{by answer\_type}} \\
    OPEN     & 645 & 0.0\,\%  & 0.0\,\%  & 0.0\,\% & 0.0\,\% & 0.0 \\
    CLOSED   & 416 & \textbf{61.5\,\%} & \textbf{61.8\,\%} & 0.0\,\% & 0.0\,\% & 0.3 \\
    \midrule
    \multicolumn{7}{l}{\emph{by content\_type}} \\
    Organ        & 253 & 55.3\,\% & 55.7\,\% & 0.0\,\% & 0.0\,\% & 0.4 \\
    Plane        &  58 & 43.1\,\% & 43.1\,\% & 0.0\,\% & 0.0\,\% & 0.0 \\
    Abnormality  & 150 & 23.3\,\% & 23.3\,\% & 0.0\,\% & 0.0\,\% & 0.0 \\
    Modality     & 108 & 21.3\,\% & 21.3\,\% & 0.0\,\% & 0.0\,\% & 0.0 \\
    KG           & 148 & 14.2\,\% & 14.2\,\% & 0.0\,\% & 0.0\,\% & 0.0 \\
    Position     & 186 &  6.5\,\% &  6.5\,\% & 0.0\,\% & 0.0\,\% & 0.0 \\
    \midrule
    \multicolumn{7}{l}{\emph{by modality}} \\
    CT     & 472 & 29.2\,\% & 29.4\,\% & 0.0\,\% & 0.0\,\% & 0.2 \\
    MRI    & 228 & 23.2\,\% & 23.2\,\% & 0.0\,\% & 0.0\,\% & 0.0 \\
    X-Ray  & 361 & 18.0\,\% & 18.0\,\% & 0.0\,\% & 0.0\,\% & 0.0 \\
    \bottomrule
  \end{tabular}
  \caption{SLAKE-En subpopulation flag rate ($\Pr[\Delta > 100]$).
  CheXagent-8b and LLaVA-OneVision-7B exhibit
  near-identical flag rates in every stratum (including the
  small \texttt{Plane} bucket, $n{=}58$) while InternVL3 and
  Qwen2.5-VL show no signal anywhere. \textbf{|gap|} is the absolute
  difference between the two high-Min-K models, in percentage
  points.}
  \label{tab:slake-subpop}
\end{table}

Table~\ref{tab:slake-subpop} and Figure~\ref{fig:slake-subpop}
locate \emph{where} the inter-cohort spread on SLAKE-En lives. The
gap is not diffuse: it is concentrated in
\texttt{answer\_type=CLOSED} (61.5\,\% vs 0\,\% on open questions),
and within closed questions it is strongest for organ identification
and plane orientation. The CheXagent and LLaVA-OneVision flag rates
match within $0.5$\,pp in every stratum we tabulate, including
strata as small as $n{=}58$ (\texttt{Plane}, $43.1\,\%$ vs
$43.1\,\%$). In an earlier draft we read this matched per-stratum
rate as a replication of a shared-exposure signature. The external
baseline in Section~\ref{sec:falsification} shows BLIP-2 produces a
numerically similar stratum profile, so we now interpret
Table~\ref{tab:slake-subpop} as a per-stratum picture of the
benchmark--cohort interaction: the CLOSED stratum is where SLAKE-En
has the lowest answer-token entropy, and that is exactly where any
high-mean-logprob VLM will sit furthest above the negative-outlier
cohort models. The InternVL3 and Qwen2.5-VL columns are
zero-everywhere because both are negative outliers on the
Min-K\%++ scale, not because their training mixes excluded SLAKE.

\section{Negative controls}
\label{sec:negcontrol}

We confirm that the image-NN threshold $\tau$ does not produce
false positives on truly out-of-domain images, with two stacked
negative controls to address two distinct null hypotheses.

\paragraph{Paintings (style-OOD).} Across 2,000 paintings
randomly sampled from \texttt{huggan/wikiart}, \emph{zero} cross the
threshold under either backbone. The minimum nearest-neighbour
cosine distance over the 2,000-painting batch is $0.143$ (B-16) and
$0.201$ (SO400M), $8\times$ and $13\times$ the respective
thresholds. This rules out the trivial null in which \emph{any}
non-medical batch would trip the detector.

\paragraph{Natural photographs (content-OOD).}
Paintings are far from PMC-OA-beta both in style (painted, not
photographic) \emph{and} in content (still lifes, landscapes,
portraits), so the painting control bounds only the joint null.
To bound the strictly stronger content-only null we re-run the
protocol (\texttt{scripts/run\_negative\_control\_natural.sbatch})
against 2,000 natural photographs sampled from
\texttt{zh-plus/tiny-imagenet}, matching the photographic
imaging regime while remaining disjoint from medical content.
\emph{Zero} of the 2,000 photographs cross the threshold under
either backbone. The minimum nearest-neighbour cosine distance is
$0.067$ (B-16) and $0.093$ (SO400M), $3.7\times$ and $6.0\times$
the respective thresholds, and the $1$st-percentile distances
($0.088$ and $0.123$) sit well clear of the flag region. The
detector therefore does not fire even on natural photographs that
share the photographic regime of medical images and differ only in
content, confirming that the SLAKE-En flag rate reflects genuine
same-view near-neighbour overlap with the PMC-OA-beta image
distribution rather than a detector that
trips on any photographic input.

Combined with the in-domain clean control (VQA-RAD,
$\sim$0.9\,\% flagged under SO400M), the two negative controls
bracket detector FPR end-to-end. Full per-quantile statistics for
the painting control are in Appendix~\ref{app:negctrl}.

\section{Recommendations and limitations}
\label{sec:limitations}

\subsection{Practical recommendations}
\label{sec:recommendations}

\begin{itemize}
\item \textbf{Use VQA-RAD as the comparatively safer benchmark} for
general open VLMs evaluated on medical questions. VQA-RAD is the
only benchmark in this audit that is clean on every cell-internal
detector (image-NN, exchangeability) for every model.
\item \textbf{Sanitize SLAKE-En for image-side source overlap before
reporting.} Image-side near-neighbour analysis flags $19.8\,\%$ of
SLAKE-En images under B-16 and $4.2\,\%$ under SO400M as having an
extreme same-view PMC-OA-beta neighbour. Even though adjudication shows
these are predominantly same-view different-patient matches rather than
exact duplicates (Section~\ref{sec:image-nn}), the strong source
overlap is itself a reason to inspect or down-weight the SO400M-flagged
subset before reporting or comparing accuracy on SLAKE-En. This
recommendation rests on Detector 1 alone and is not affected by
the falsification in Section~\ref{sec:falsification}.
\item \textbf{Treat one held-out exchangeability hit as actionable
and treat the second as a release-order artefact.}
Qwen2.5-VL\,$\times$\,SLAKE-En survives every robustness check we
applied: refined exchangeability $p = 5.0 \times 10^{-4}$ at
$10\,000$ permutations (Bonferroni-significant at $\alpha=0.05$
over the 27-cell audit grid, $p_{\text{Bonf}}=0.0135$),
hash-reordered $p=0.68$ (signal
disappears under content reordering), and \emph{both} external
non-medical baselines stay null on SLAKE-En release (BLIP-2
$p=0.31$, InstructBLIP $p=0.97$). This cell is the strongest
held-out memorization candidate in the study.
LLaVA-OneVision\,$\times$\,PathVQA does \emph{not} survive: BLIP-2
fires PathVQA release exchangeability at the same magnitude
($p = 1.6 \times 10^{-3}$ refined; hash $p=0.28$), while a second
baseline, InstructBLIP, is null ($p=1.000$). Because BLIP-2
cannot have medical-VQA training exposure, the shared release-only
signature implicates a release-order property of PathVQA itself,
consistent with the benchmark being shipped grouped by source
slide, and not LLaVA-OneVision-specific memorization. The cell
also falls below Bonferroni significance at $\alpha=0.05$ on the
27-cell grid ($p_{\text{Bonf}}=0.054$). We
retract the LLaVA-OneVision\,$\times$\,PathVQA actionable flag
from earlier drafts of this audit.
Exchangeability remains cell-internal and not subject to the
cohort-calibration confound, but the ordering ablation
(Section~\ref{sec:order-ablation}) is a required follow-up before
treating any held-out hit as memorization-specific evidence.
\item \textbf{Treat the OmniMedVQA mirror as contaminated for
text-side evaluation.} On the public 4,999-example OmniMedVQA mirror,
exchangeability fires for five medical and general models while the
external BLIP-2 baseline remains clean (Section~\ref{sec:omnimedvqa}).
Because
the released mirror exposes only a \texttt{train} split, it should not
be used as a contamination-sensitive held-out benchmark for open
medical VLMs.
\item \textbf{Do not rely on cohort-relative Min-K\%++ tail
enrichment or cross-model top-$K$ overlap as standalone
membership-inference signals} on small ($\leq 5$-model) medical-VLM
cohorts. Both signals are confounded by inter-model calibration
heterogeneity (Section~\ref{sec:falsification}). If used, they
must be paired with an external pre-medical baseline; a signal
that reproduces in the baseline must be re-classified as
benchmark-text predictability rather than memorization.
\end{itemize}

\subsection{Limitations}

\paragraph{Cohort calibration heterogeneity (tested and falsified).}
Detectors 3 and 4 are cohort-relative: they compare a target
model's Min-K\%++ scores against the median (or top-$K$ set) of
the other cohort models. This is only valid if the cohort is
calibrated homogeneously on the surface forms in the benchmark.
The four-model cohort used in the headline scan is not: Qwen2.5-VL
and InternVL3 are systematic negative outliers on SLAKE-En, which
pulls the median down and inflates $\Delta_i$ uniformly for the
three remaining models. We tested this directly in
Section~\ref{sec:falsification} by extending the cohort with
BLIP-2, a pre-medical baseline that cannot share medical-VQA
exposure. BLIP-2 reproduces the tail-enrichment flag
($\Pr[\Delta>100]=19.79\,\%$, $\Delta_{\max}=1703$) and the
top-25 set on SLAKE-En verbatim. We accordingly reclassify the
original SLAKE-En cohort-relative findings as a
benchmark-cohort-interaction artefact rather than memorization
evidence, and we recommend any future use of these detectors on
medical-VLM cohorts be paired with an external pre-domain
baseline. The subpopulation breakdown of
Section~\ref{sec:subpop} is a description of \emph{where} the
inter-cohort gap lives (concentrated in low-entropy CLOSED items),
not evidence that those items were memorized.

\paragraph{What the falsification does not touch.} The image-side
near-neighbour findings (Detector 1) and the cell-internal
canonical-order exchangeability tests (Detector 2) do not depend on
cohort-relative calibration. In particular, the
SLAKE-En\,$\times$\,Qwen2.5-VL survivor is unaffected by the BLIP-2
cohort-calibration falsification.

\paragraph{Coverage.} We audit four open VLMs in the primary
held-out scan, and we additionally include two external non-medical
baselines (BLIP-2 and InstructBLIP) for the falsification and
ordering-ablation analyses and the gated medical model MedGemma-4B-IT
in the cell-internal exchangeability family, on three held-out
benchmarks plus an auxiliary OmniMedVQA text-side extension. Several
further medical VLMs (RadFM, LLaVA-Med, Med-Flamingo, MedDr) and
benchmarks (MIMIC-CXR-VQA, CheXpert, EchoNet-Dynamic-VQA) are left to
future work. Our OmniMedVQA audit is text-side only; an image-side
near-neighbour scan of OmniMedVQA against PMC-OA-beta would extend the
coverage of Detector~1 and is a natural next step.

\paragraph{Causal inference.} Our findings identify statistical
contamination signals, not provable training-data inclusion. A
positive signal is strong circumstantial evidence; ruling out
training exposure requires access we do not have.

\paragraph{Permutation cost.} We sample 10\,000 permutations per
exchangeability test, giving a smallest reportable $p$-value of
$\approx 10^{-4}$. Because the per-example $\sum \log p$ values are
saved per cell, we can refine any cell to a tighter floor offline
without re-running the model: a fresh permutation draw needs only the
saved scalars and takes $\sim$1\,minute per cell on a single CPU. The
Bonferroni-corrected threshold at family-wise $\alpha=0.01$
($3.7{\times}10^{-4}$ for our 27-cell grid) sits strictly above the
$10^{-4}$ floor, so any floor-pinned cell is automatically
Bonferroni-significant; cells reported here pass both raw and
multiplicity-corrected significance unless noted.

\impact{%
This work is a data-quality audit: it asks whether widely used public
medical visual-question-answering benchmarks share enough provenance
with web-scale pretraining corpora to compromise their use as held-out
evaluations. We see three positive impacts. First, the audit gives the
medical-ML community a reusable, calibrated procedure for stating what
a benchmark number does and does not control for, which supports more
trustworthy reporting of clinical-model accuracy. Second, by insisting
on external pre-domain falsification and by manually adjudicating the
image-side flags, we deliberately push against the failure mode of
over-claiming ``contamination'' or ``leakage'' from a single
unvalidated detector, a failure mode that can unfairly discredit
honest benchmarks and models. Third, all benchmarks we audit are
already public, and we introduce no new patient data, so the audit
adds no privacy exposure beyond what the source corpora already carry.

We also note risks and limitations of dual-use type. The same
nearest-neighbour and membership-style detectors that flag benchmark
overlap could, in principle, be aimed at inferring whether a specific
clinical record was used to train a model; we therefore frame the
detectors as cohort-level, benchmark-integrity tools and report their
substantial false-positive behaviour in-domain (Section~\ref{sec:negcontrol},
Appendix~\ref{app:confound}) precisely so they are not mistaken for
reliable per-record membership oracles. A second risk is
mis-interpretation: a positive overlap signal is statistical
circumstantial evidence, not proof of training-set inclusion, and
should not be used to make definitive claims about a specific model's
training data or to penalise a benchmark without the kind of
adjudication we demonstrate here. We surface these caveats throughout
the paper and in the claims-to-evidence map (Appendix~\ref{app:claims})
so that downstream users inherit the appropriate level of caution.%
}

\section*{Data and code availability}
All code and derived data needed to reproduce the audit are released at
\url{https://github.com/brucechanglongxu/medvlm-contamination-audit}.
The release includes the flagged SLAKE-En example and image IDs with
their PMC-OA-beta nearest-neighbour IDs and cosine distances, the
manual adjudication manifest underlying Table~\ref{tab:adjudication},
the saved per-example exchangeability log-likelihood scalars and the
permutation seeds, the per-example gold-answer log-probabilities and
source-overlap flags behind the SLAKE-En inflation estimate
(\texttt{scripts/overlap\_inflation.py}), and the driver scripts for the
threshold sweep
(\texttt{scripts/tau\_sensitivity.py}), the synthetic confound
demonstration (\texttt{scripts/confound\_simulation.py}), and the
BLIP-2 external-baseline falsification. The audited models and
benchmarks are the public Hugging Face releases listed in
Appendix~\ref{app:repro}; no new data were collected from human
subjects.

\bibliography{refs}

\appendix

\section{Negative-control details}
\label{app:negctrl}

\begin{table}[ht]
  \centering
  \begin{tabular}{lcccccc}
    \toprule
    Backbone & $n_{\text{flag}}/n$ & $\min$ & $q_{0.01}$ & $q_{0.05}$ & $q_{0.50}$ & $\tau$ \\
    \midrule
    SigLIP-B-16    & 0/2000 & 0.143 & 0.182 & 0.216 & 0.305 & 0.0180 \\
    SigLIP-SO400M  & 0/2000 & 0.201 & 0.265 & 0.308 & 0.407 & 0.0154 \\
    \bottomrule
  \end{tabular}
  \caption{Wikiart paintings vs PMC-OA-beta.
  Seed 0; queries drawn uniformly from the
  first $\max(4n, n+50)$ rows of \texttt{huggan/wikiart::train}
  under streaming.}
\end{table}

\section{Synthetic demonstration of the cohort-median confound}
\label{app:confound}

The external-baseline falsification of
Section~\ref{sec:falsification} is reproduced here from first
principles, with no model and no GPU, by instantiating the
calibration model of Equation~\eqref{eq:calib} as a deterministic
synthetic audit. We draw a benchmark of $n=1{,}061$ examples (the
SLAKE-En test size) whose easiness $e_i$ follows a two-component
structure matching SLAKE-En: a $60\,\%$ OPEN stratum with modest
easiness ($e_i = |\mathcal{N}(0,1)|$) and a $40\,\%$ CLOSED stratum
with a heavy right tail ($e_i \sim \mathrm{Exp}(900\text{ nats})$),
mirroring the low-entropy closed-form items that dominate the
observed $\Delta$ tail. We assign each model a calibration gain
$g_m$ and offset $b_m$ but set the contamination term
$\delta_{i,m}\equiv 0$ \emph{everywhere}: no model has seen any
example. The cohort comprises two low-gain negative-outlier models
($g=0.05$), two high-gain models ($g=1.0$), and a third high-gain
model standing in for the provably clean external baseline
($g=1.0$).

\begin{figure}[ht]
  \centering
  \includegraphics[width=\linewidth]{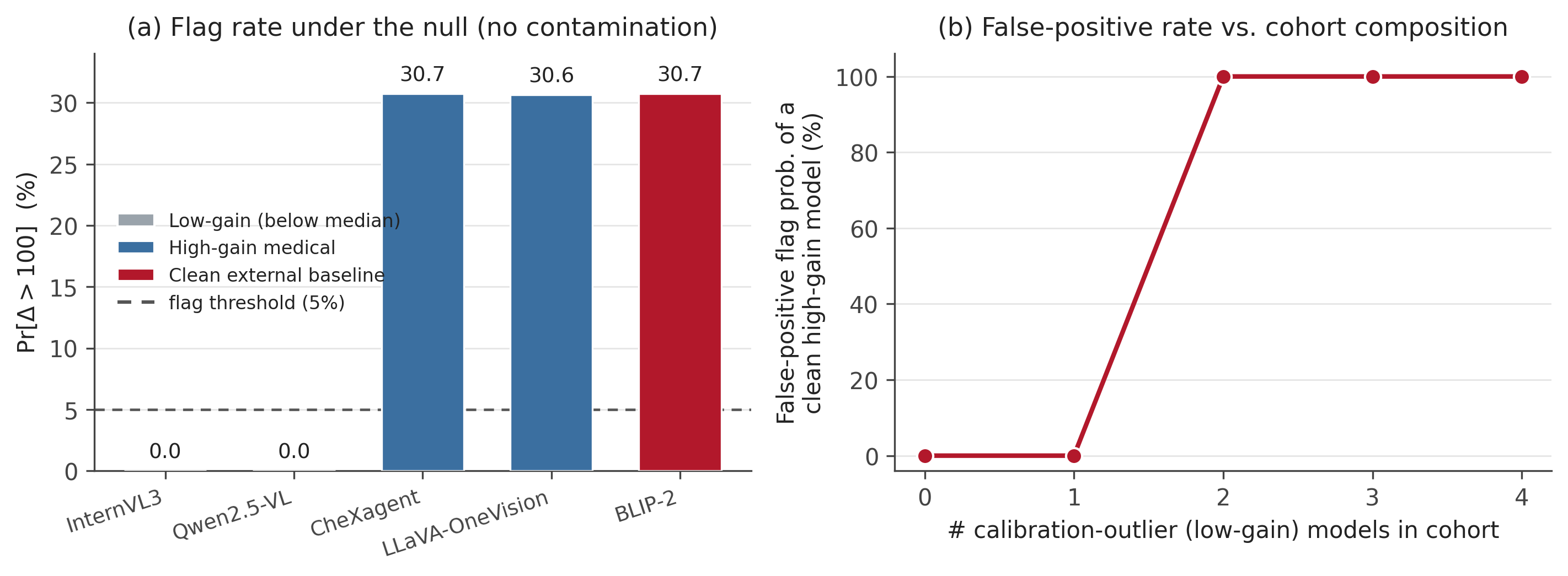}
  \caption{Synthetic audit under the calibration model of
  Equation~\eqref{eq:calib} with \emph{no contamination anywhere}.
  \textbf{(a)} Detector~3 tail-enrichment rate $\Pr[\Delta>100]$ per
  model: the two low-gain calibration outliers (grey) register zero,
  while every above-median-gain model fires, including the clean
  external baseline (red) at parity with the two ``suspect'' high-gain
  models (blue). \textbf{(b)} False-positive flag probability of a
  single clean high-gain probe as a function of the number of low-gain
  outliers in the cohort: the probe is never flagged until the cohort
  median collapses onto the outliers (at two outliers), after which it
  is flagged in every draw. Both panels are produced by
  \texttt{scripts/confound\_simulation.py} under a fixed seed.}
  \label{fig:confound-sim}
\end{figure}

Figure~\ref{fig:confound-sim}(a) shows the result of scoring this
contamination-free cohort with Detector~3. Both low-gain models
register $\Pr[\Delta>100]=0$, while \emph{all three} high-gain
models, including the clean baseline, fire the tail-enrichment flag
at $\Pr[\Delta>100]=30.7\,\%$ with $\Delta_{\max}\approx 3{,}400$,
qualitatively reproducing the empirical SLAKE-En tail: the synthetic
$\Pr[\Delta>100]=30.7\,\%$ is the same order of magnitude as the
empirical $\sim$24\,\%, while $\Delta_{\max}$ matches closely
(Table~\ref{tab:falsification}). Figure~\ref{fig:confound-sim}(b)
sweeps cohort composition: holding two high-gain anchor models
fixed and adding a single clean high-gain probe, the probe's
false-positive flag probability is $0$ until the cohort contains
\emph{two} low-gain outliers, at which point the cohort median
collapses onto the outliers, $\Gamma_{\text{probe}}>0$, and the
clean probe is flagged in $100\,\%$ of draws. This is the precise
mechanism by which the four-model SLAKE-En cohort manufactured the
original Detector~3/4 signal, and it confirms that the only safe
remedy is an external pre-domain baseline rather than a larger
in-domain cohort. The simulation is deterministic under a fixed
seed; the driver is
\texttt{scripts/confound\_simulation.py}.

\section{Image near-neighbour adjudication}
\label{app:adjudication}

Table~\ref{tab:adjudication} lists every distinct SLAKE-En image
flagged by the image nearest-neighbour detector
(Section~\ref{sec:image-nn}) under either SigLIP backbone, ordered by
strictest cosine distance, together with the matched PMC-OA-beta figure
and a per-pair visual verdict. We rendered a side-by-side panel for each
pair (\texttt{scripts/adjudication\_pairs.py}, output in
\texttt{outputs/adjudication/}) and inspected it. Every flagged pair is
the same imaging modality and projection, but only one (example~12504)
is an ambiguous candidate for an exact pixel-level duplicate; the
remaining nineteen are same-view images of different patients. The
recurrence of individual PMC figures as the nearest neighbour of
multiple SLAKE images (\texttt{PMC2569031}, \texttt{PMC9062550}) further
indicates that the detector responds to a shared same-view image
distribution rather than to unique memorized duplicates. We report the
verdicts as conservative author-side adjudication; they are not a
substitute for radiologist review or exact-match provenance.

{\footnotesize\setlength{\tabcolsep}{3pt}
\begin{longtable}{llllcl}
\caption{Manual adjudication of all 20 distinct SLAKE-En images flagged
by the image nearest-neighbour detector under either SigLIP backbone,
ordered by strictest (smallest) cosine distance. Distances are to the
nearest PMC-OA-beta figure. Every flagged pair is the same imaging
modality and projection; the verdict records whether visual inspection
of the rendered panel (\texttt{outputs/adjudication/}) supports an exact
duplicate. Exact pixel-duplication is supported in at most one case; the
remaining pairs are same-view images of different patients.}
\label{tab:adjudication}\\
\toprule
SLAKE ex. & SLAKE img & PMC article & PMC figure & cos-dist & verdict \\
\midrule
\endfirsthead
\toprule SLAKE ex. & SLAKE img & PMC article & PMC figure & cos-dist & verdict \\ \midrule \endhead
12474 & \texttt{xmlab375} & PMC6195915 & \texttt{PMC6195915\_fig1$\dots$} & 0.0124 & same-view, diff.\ patient \\
12504 & \texttt{xmlab385} & PMC9062550 & \texttt{PMC9062550\_fig-0001$\dots$} & 0.0128 & ambiguous (poss.\ duplicate) \\
12369 & \texttt{xmlab338} & PMC6684717 & \texttt{PMC6684717\_Fig13$\dots$} & 0.0128 & same-view, diff.\ patient \\
11994 & \texttt{xmlab142} & PMC1277010 & \texttt{PMC1277010\_F1\_3739} & 0.0131 & same-view, diff.\ patient \\
12384 & \texttt{xmlab344} & PMC9062550 & \texttt{PMC9062550\_fig-0001$\dots$} & 0.0134 & same-view, diff.\ patient \\
12961 & \texttt{xmlab82} & PMC5728928 & \texttt{PMC5728928\_F1\_252744} & 0.0137 & same-view, diff.\ patient \\
12489 & \texttt{xmlab378} & PMC9021672 & \texttt{PMC9021672\_f5\_257345} & 0.0144 & same-view, diff.\ patient \\
12031 & \texttt{xmlab160} & PMC2531090 & \texttt{PMC2531090\_F1\_27449} & 0.0149 & same-view, diff.\ patient \\
12519 & \texttt{xmlab386} & PMC2569031 & \texttt{PMC2569031\_F2\_28934} & 0.0151 & same-view, diff.\ patient \\
12339 & \texttt{xmlab320} & PMC4235038 & \texttt{PMC4235038\_F1\_337068} & 0.0153 & same-view, diff.\ patient \\
12459 & \texttt{xmlab374} & PMC8424947 & \texttt{PMC8424947\_Fig3$\dots$} & 0.0155 & same-view, diff.\ patient \\
12023 & \texttt{xmlab159} & PMC8229150 & \texttt{PMC8229150\_f004$\dots$} & 0.0158 & same-view, diff.\ patient \\
12249 & \texttt{xmlab274} & PMC7953744 & \texttt{PMC7953744\_Fig1$\dots$} & 0.0159 & same-view, diff.\ patient \\
12399 & \texttt{xmlab350} & PMC6027734 & \texttt{PMC6027734\_Fig2$\dots$} & 0.0163 & same-view, diff.\ patient \\
12106 & \texttt{xmlab224} & PMC8992766 & \texttt{PMC8992766\_F1\_247118} & 0.0164 & same-view, diff.\ patient \\
12207 & \texttt{xmlab253} & PMC8686661 & \texttt{PMC8686661\_Fig1$\dots$} & 0.0165 & same-view, diff.\ patient \\
12822 & \texttt{xmlab557} & PMC4302363 & \texttt{PMC4302363\_fig1$\dots$} & 0.0168 & same-view, diff.\ patient \\
12834 & \texttt{xmlab561} & PMC5815738 & \texttt{PMC5815738\_F3\_276750} & 0.0177 & same-view, diff.\ patient \\
12309 & \texttt{xmlab312} & PMC2569031 & \texttt{PMC2569031\_F2\_28934} & 0.0179 & same-view, diff.\ patient \\
11987 & \texttt{xmlab135} & PMC8103267 & \texttt{PMC8103267\_f001$\dots$} & 0.0180 & same-view, diff.\ patient \\
\bottomrule
\end{longtable}
}

\section{Reproducibility}
\label{app:repro}

We summarize the information needed to reproduce every number, table,
and figure in this paper. All analyses are deterministic given the
models, benchmarks, and seeds below; no result depends on
unseeded randomness.

{\raggedright \paragraph{Models and benchmarks.} The audited VLMs are the public
Hugging Face checkpoints \texttt{OpenGVLab/\allowbreak InternVL3-8B},
\texttt{Qwen/\allowbreak Qwen2.5-\allowbreak VL-7B-\allowbreak Instruct},
\texttt{StanfordAIMI/\allowbreak CheXagent-\allowbreak 8b},
\texttt{llava-hf/\allowbreak llava-onevision-\allowbreak qwen2-7b-ov-hf}, with
\texttt{google/\allowbreak medgemma-4b-it} (gated) in the OmniMedVQA text-side
family and \texttt{Salesforce/\allowbreak blip2-opt-2.7b} and
\texttt{Salesforce/\allowbreak instructblip-vicuna-7b} as external pre-2023
baselines. Benchmarks are the public releases of SLAKE-En
\citep{slake2021}, PathVQA \citep{pathvqa2020}, VQA-RAD
\citep{vqarad2018}, and a 4{,}999-example public mirror of OmniMedVQA
\citep{omnimedvqa2024}; the image-side corpus is PMC-OA-beta
\citep{pmcoa2023} (1.85\,M figures).\par}

\paragraph{Detector parameters.} Detector~1 (image near-neighbour)
embeds images with OpenCLIP \texttt{ViT-B-16-SigLIP} and
\texttt{ViT-SO400M-14-SigLIP}; a benchmark image is flagged when its
cosine distance to its PMC-OA-beta nearest neighbour falls below
$\tau=Q_\alpha$, the $\alpha=0.01$ quantile of a $5{,}000$-sample null
of intra-corpus natural-image distances (thresholds
$\tau_{\text{B-16}}=0.0180$, $\tau_{\text{SO400M}}=0.0154$;
Appendix figure~\ref{fig:tau-sensitivity} sweeps $\alpha$).
Detector~2 (canonical-order exchangeability) uses $10{,}000$
permutations, giving a smallest reportable $p$-value of
$\approx 10^{-4}$; the family-wise Bonferroni threshold over the
$m=27$ held-out cells at $\alpha=0.01$ is $3.7{\times}10^{-4}$.
Detector~3 (cohort-relative Min-K\%++ tail enrichment) uses $K=20\%$
of lowest-$z$ tokens per example and flags a cell when
$\Pr[\Delta_i>100]>5\%$; Detector~4 (cross-model top-$K$ overlap)
uses $K=25$. Detectors~3 and~4 operate on the four-model medical
cohort plus the BLIP-2 external baseline used in the falsification.

\paragraph{Scripts and seeds.} The synthetic confound demonstration
(Appendix~\ref{app:confound}, Figure~\ref{fig:confound-sim}) is
produced by \texttt{scripts/confound\_simulation.py} under fixed seed
$20260531$. The threshold-sensitivity figure
(Figure~\ref{fig:tau-sensitivity}) is re-plotted from committed
per-benchmark sweep data by \texttt{scripts/tau\_sensitivity.py}. The
image near-neighbour adjudication panels and the manifest underlying
Table~\ref{tab:adjudication} are generated by
\texttt{scripts/adjudication\_pairs.py}. Exchangeability $p$-values are
refinable offline from the saved per-example $\sum\log p$ scalars
without re-running any model: a fresh permutation draw needs only the
saved scalars and takes $\sim$1\,minute per cell on a single CPU.

\paragraph{Compute.} The model-likelihood scans run on a single
80\,GB GPU per (model, benchmark) pair; the image-embedding and
nearest-neighbour passes run on a single GPU plus CPU for the FAISS
index over PMC-OA-beta. The synthetic, sensitivity, and adjudication
analyses in the appendices require no GPU.

\section{Claims-to-evidence map}
\label{app:claims}

Table~\ref{tab:claims} maps each claim in the paper to its supporting
evidence and its status under control. We distinguish claims that
\emph{survive} an external pre-domain baseline or ordering ablation
from detectors that \emph{collapse} once such a control is added.

{\small
\begin{longtable}{p{0.46\linewidth} l p{0.30\linewidth}}
\caption{Each claim, its status under control, and where it is
established.}
\label{tab:claims}\\
\toprule
\textbf{Claim} & \textbf{Status} & \textbf{Evidence} \\
\midrule
\endfirsthead
\toprule
\textbf{Claim} & \textbf{Status} & \textbf{Evidence} \\
\midrule
\endhead
\bottomrule
\endfoot
SLAKE-En shows extreme same-view image near-neighbour overlap with
PMC-OA-beta ($19.8\%$ B-16, $4.2\%$ SO400M) vs.\ $\leq0.9\%$ for
in-domain VQA-RAD and $0\%$ for two out-of-domain controls &
survives &
\S\ref{sec:image-nn}, Fig.~\ref{fig:imgnn-flagrate} \\
The image signal is robust to the threshold $\tau$ across two orders
of magnitude of $\alpha$ &
survives &
Fig.~\ref{fig:tau-sensitivity} \\
The flagged image pairs are same-modality, same-projection matches to
\emph{different} patients ($\leq1$ exact-duplicate candidate of $20$);
the signal is source/distributional overlap, not per-image
memorization &
qualified &
\S\ref{sec:image-nn}, App.~\ref{app:adjudication},
Table~\ref{tab:adjudication} \\
OmniMedVQA canonical-order exchangeability fires for all five medical
and general VLMs while the non-medical baseline BLIP-2 stays clean &
survives &
\S\ref{sec:omnimedvqa} \\
The SLAKE-En\,$\times$\,Qwen2.5-VL exchangeability hit survives an
ordering ablation and two external baselines &
survives &
\S\ref{sec:regimes}, \S\ref{sec:order-ablation} \\
The PathVQA exchangeability hit is a release-order artefact
(reproduced by a non-medical baseline; vanishes under content-derived
reordering) &
reattributed &
\S\ref{sec:pathvqa-llava}, \S\ref{sec:order-ablation} \\
Cohort-relative Min-K\%++ tail enrichment (Detector~3) and cross-model
top-$K$ overlap (Detector~4) collapse once BLIP-2 is added to the
cohort: a model that cannot share medical-VQA exposure is flagged
identically &
collapses &
\S\ref{sec:falsification} \\
The Detector~3/4 confound is a general consequence of inter-model
calibration heterogeneity (gain-gap), not a quirk of our cohort &
collapses &
\S\ref{sec:falsification} (Eq.~\ref{eq:delta-decomp}),
App.~\ref{app:confound}, Fig.~\ref{fig:confound-sim} \\
VQA-RAD is clean on both the image- and text-side signals and serves
as the per-benchmark sanity check &
clean control &
\S\ref{sec:regimes}, \S\ref{sec:image-nn} \\
\end{longtable}
}

\end{document}